\documentclass[journal,onecolumn]{IEEEtran}
\ifCLASSINFOpdf
\else
\fi
\usepackage{amsmath}
\usepackage{color}

\usepackage{subfigure}
\usepackage{graphicx}
\usepackage{float}
\usepackage{caption}
\usepackage{multirow}
\usepackage{hyperref}

\hypersetup{
    colorlinks=true,
    linkcolor=blue,
    filecolor=blue,
    citecolor=blue,
    urlcolor=blue,
}

\newcommand{\norm}[1]{\Big\lVert #1 \Big\rVert}

\newcommand{\argmin}{\mathop{\rm arg~min}\limits}
\newcommand{\fsum}{\mathop{\rm \sum}\limits}

\begin{document}
%
% paper title
% Titles are generally capitalized except for words such as a, an, and, as,
% at, but, by, for, in, nor, of, on, or, the, to and up, which are usually
% not capitalized unless they are the first or last word of the title.
% Linebreaks \\ can be used within to get better formatting as desired.
% Do not put math or special symbols in the title.
\title{\huge Efficient Consensus Model based on Proximal Gradient Method applied to Convolutional Sparse Problems}
%
%
% author names and IEEE memberships
% note positions of commas and nonbreaking spaces ( ~ ) LaTeX will not break
% a structure at a ~ so this keeps an author's name from being broken across
% two lines.
% use \thanks{} to gain access to the first footnote area
% a separate \thanks must be used for each paragraph as LaTeX2e's \thanks
% was not built to handle multiple paragraphs
%

\author{Gustavo~Silva
        and~Paul~Rodriguez% <-this % stops a space
\thanks{G. Silva and P. Rodriguez are with the Department of
Electrical Engineering, Pontificia Universidad Cat\'olica del Per\'u, Lima, Per\'u. E-mail : gustavo.silva@pucp.edu.pe and prodrig@pucp.edu.pe}% <-this % 
}

\maketitle

% As a general rule, do not put math, special symbols or citations
% in the abstract or keywords.
\begin{abstract}
Convolutional sparse representation (CSR), shift-invariant model for inverse problems, has gained much attention in the fields of signal/image processing, machine learning and computer vision. The most challenging problems in CSR implies the minimization of a composite function of the form $min_x\sum_i f_i(x) + g(x)$, where a direct and low-cost solution can be difficult to achieve. However, it has been reported that semi-distributed formulations such as ADMM consensus can provide important computational benefits.

In the present work, we derive and detail a thorough theoretical analysis of an efficient consensus algorithm based on proximal gradient (PG) approach. The effectiveness of the proposed algorithm with respect to its ADMM counterpart is primarily assessed  in the classic convolutional dictionary learning problem. 
Furthermore, our consensus method, which is generically structured, can be used to solve other optimization problems, where  a sum of convex functions with a regularization term share a single global variable. As an example, the proposed algorithm is also applied to another particular convolutional problem for the anomaly detection task.
\end{abstract}

% Note that keywords are not normally used for peerreview papers.

% For peer review papers, you can put extra information on the cover
% page as needed:
% \ifCLASSOPTIONpeerreview
% \begin{center} \bfseries EDICS Category: 3-BBND \end{center}
% \fi
%
% For peerreview papers, this IEEEtran command inserts a page break and
% creates the second title. It will be ignored for other modes.
\IEEEpeerreviewmaketitle

\section{\bfseries Introduction}

\IEEEPARstart{S}{parse} representation (SR) is as an effective technique for solving underdetermined linear system of  equations. Under a sparsity constraint, the SR model assumes that a sectorized signal $s \in  {\rm I\!R}^N$  can be approximated as a linear combination of a few columns (atoms) of an overcomplete matrix $D \in  {\rm I\!R}^{N\times M}$ (dictionary), i.e. $s \approx Dx$, where $x\in  {\rm I\!R}^{M}$ is a sparse. Unfortunately, its base structure requires the atoms to be perfectly aligned with the analyzed signal vectors or image patches, resulting in a non-translational invariant approach. When learning a dictionary $D$ to encodes a collection of images via overlapping patches, many of the atoms are also shifted versions of each other, generating a higher computational burden due to redundant information. 

Convolutional sparse representation (CSR) has been designed to overcome the aforementioned drawbacks by modeling entire signals or images as a sum of convolutions between dictionary filters $d_m$ and coefficient maps $x_m$.  
In particular, the CSR model encloses two optimization problems: Convolutional Dictionary Learning (CDL) and Convolutional Sparse Coding (CSC), which have been extensively and successfully used in a variety of applications such as classification \cite{zhou2014classification}, 
denoising \cite{li2018video,simon2019rethinking}, anomaly detection  \cite{carrera2015detecting,pilastre2020anomaly}, super-resolution \cite{gu2015convolutional} and  more. In  the  literature,  many  investigations  that  have been outlined in this convolutional framework aim to reduce the computational expenses produced by solving  a linear  system of convolutions $s\approx \sum_m d_m*x_m$. The most popular methodology \cite{bristow2013fast,wohlberg2014efficient,wohlberg2015efficient,vsorel2016fast,garcia-2017-subproblem,garcia2018convolutional,silva2018efficient} consists in 
 performing the inversion of the convolutional system in the frequency domain.
Recently, the spatial domain has received a renewed interest by slice based algorithms \cite{papyan2017convolutional,simon2019rethinking,rey2020variations}, which allow capturing and enhancing local features like its patch based predecessor.

Independent of the methodology, the most challenging optimization problem identified in the CSR, described in general form, is:
\begin{equation}
\argmin _{\{ {z}\}}~
F(z) =
\fsum_{i=1}^{R} f_i({z}) +  g({z}). 
\label{eq:dist_reg}
\end{equation}
where a common variable $z$ and global regularization term makes difficult to get a direct and efficient solution.

When the problem (\ref{eq:dist_reg}) does not include a global regularization term $g(z)$  or a global constraint, on the field of distributed optimization, it has been proposed many decentralized algorithms  \cite{shi2015extra,yuan2016convergence,zeng2018nonconvex,reisizadeh2019exact} that minimize locally each function $f_i(z)$ with reduced intercommunications.
Although the newest decentralized algorithms\footnote{ We do not consider the decentralized approach \cite{alghunaim2019linearly}, \cite{xu2020distributed} in this work as, to the best of our knowledge, there is no existing decentralized algorithm that has been applied to the CSR problems.} can fully address the problem (\ref{eq:dist_reg}), these are  oriented to  network computing applications in which are required the least data transfer  and the highest data privacy. On the CSR context, consensus formulations such as ADMM consensus \cite{garcia-2017-subproblem,choudhury2017consensus,garcia2018convolutional,pilastre2020anomaly} have shown to be the leading approach to solve (\ref{eq:dist_reg}) with low processing time, especially if  parallel deployments are involved.

In comparison to our previous work \cite{silva2018efficient_icip} focused on fast CDL algorithms, in which we introduced the APG consensus algorithm by forcing a consensus structure in a proximal gradient solution, the contributions of the present manuscript can be summarized as follows:
\begin{itemize}
    \item We elaborate the formal theoretical derivation of the APG consensus algorithm, which can be used to efficiently deal with optimization problems as (\ref{eq:dist_reg}). 
    \item  We provide full details of how to estimate the associated step-size in order to have a self-adjusting algorithm.
    \item We consider an exhaustive experimental analysis of the proposed method, not only limited to processing time, in distinct convolutional sparse problems. 
\end{itemize}

The organization of this work is as follows: Section \ref{Pre-concepts} briefly summarizes general concepts necessary to understand the proposed consensus method, the CDL framework and the CDL algorithms, consequently presented in Sections \ref{sect:PGcons} to \ref{sec:Dictup}. In Section \ref{Sect:results}, we perform a thorough evaluation of our algorithm. Finally, conclusions are reported in Section \ref{Sect:conclusions}.
\vspace{2mm}

\section{\bfseries Preliminary concepts}
\label{Pre-concepts}

Suppose we require to minimize the composite model:
\begin{equation}
\min_{\{{x}\}} \text{\hspace{2.5mm}}
F(x): =f({x}) + g({x}) 
\label{eq:fun1}
\end{equation}
where $f: {\rm I\!R}^N \to {\rm I\!R} $ is a smooth convex function
with a gradient $\nabla f$ that is $L$-Lipschitz continuous: $|| \nabla f(x) - \nabla f(y)|| \leq L(f) ||x -y||$ for some $L \geq 0$, and $g: {\rm I\!R}^N \to {\rm I\!R} $
is a possibly non-smooth convex function, which $g$'proximal operator
\begin{equation}
    prox_g(y) = \argmin_{x} \frac{1}{2}||x - y|| + g(x)
\end{equation}
has a computationally simple or affordable solution.
It is straightforward to notice that several types of algorithms (first and second-order methods \cite{beck2017first,battiti1992first}, iteratively reweighted least square \cite{lai2013improved} and ADMM \cite{boyd2011distributed}) can be
used to deal with (\ref{eq:fun1}). However, we will mainly  focus on Alternating Direction Method of Multipliers and Accelerated Proximal Gradient, two meaningful algorithms applied to CSR problems.

\subsection{\bfseries Alternating Direction Method of Multipliers}
\label{Ref-ADMM-par}
\color{black} ADMM  \cite[Ch.~2]{boyd2011distributed} is a versatile algorithm characterized by its separability property and good convergence. It can be employed to solve
\begin{equation}
\min_{\{{x}\}, \{{y}\}} \text{\hspace{2.5mm}}
f({x}) + g({y}) \text{\hspace{5mm}} s.t. \text{\hspace{5mm}} A{x} + B {y} - c = 0,
\label{eq:ADMM}
\end{equation}
where the two set of variables, ${x}$ and ${y}$, are linearly related. The ADMM iterations in scaled form are given by
\begin{eqnarray}
 {x}^{(k+1)} &=& \min_{\{{x}\}} 
f({x})  + \frac {\rho }{2}  \|{A}  {x} + {B} {y}^{(k)} -c + 
{u}^{(k)} \|_{2}^{2}   \label{eqn:admm-1} \\
 {y}^{(k+1)} &=& \min_{\{{y}\}} g({y}) 
+ \frac {\rho }{2} \| {A} {x}^{(k+1)} + {B}	 {y} - c +  {u}^{(k)} \| _{2}^{2}  \label{eqn:admm-2} \\
 {u}^{(k+1)} &=&  {u}^{(k)} +  {A} {x}^{(k+1)} +  {B} {y}^{(k+1)}-c.  \label{eqn:admm-3} 
 \end{eqnarray}
where penalty parameter $ \rho$, referred to as augmented lagrangian parameter,
controls the convergence of the algorithm.
 Most well-known variants developed to enhance the practical convergence are:  (i) Over-relaxation method  \cite[Ch.~3]{boyd2011distributed} that involves to replace $Ax^{(k+1)}$ in (\ref{eqn:admm-2}) and (\ref{eqn:admm-3}) by $\beta x^{(k+1)} - (1-\beta) (By^{(k)}-c)$, where $\beta \in (0,2]$ is the relaxation parameter. (ii) Update rule for the penalty parameter that allows to automatically vary this parameter and makes  the performance less dependent on the initial selected value. Furthermore, it can be proved \cite{chambolle2016introduction} that the ADMM algorithm can achieve a quadratic convergence rate if the function $f$ is strongly convex.

When the optimization problem (\ref{eq:fun1}) is a distributed case, e.g. $f({x}) = \sum_i f_i({x})$, as in (\ref{eq:dist_reg}), using local variables $x_i$ and a common global variable $y$, it  can be rewritten as a
global consensus problem \cite[Ch.~7]{boyd2011distributed}  with a regularization term:
\begin{equation}
\min_{\{{x}_i\}, \{{y}\}} \text{\hspace{2.5mm}}
 \fsum_{i=1}^{R} f_i({x_i}) + g({y}) \text{\hspace{5mm}} s.t. \text{\hspace{5mm}} {x_i} = {y}.
\label{eq:ADMM}
\end{equation}
Now the scaled ADMM algorithm, called ADMM Consensus algorithm, would be
\begin{eqnarray}
 {x}_i^{(k+1)} &=& \min_{\{{x}\}} 
f_i({x}_i)  + \frac {\rho }{2}  \| {x}_i +  {y}^{(k)}  + 
{u}_i^{(k)} \|_{2}^{2}   \label{eqn:admm-cns-1} \\
 {y}^{(k+1)} &=& \min_{\{{y}\}} 
g({y}) 
+ \frac {\rho }{2} \fsum_{i=1}^{R} \| {x}_i^{(k+1)} + {y}  +  {u}_i^{(k)} \| _{2}^{2}  \label{eqn:admm-cns-2} \\
 {u}^{(k+1)} &=& {u}_i^{(k)} +  {x}_i^{(k+1)} +   {y}^{(k+1)}.  \label{eqn:admm-cns-3}
\end{eqnarray}
If we collect the linear and quadratic terms, the $y$-update (\ref{eqn:admm-cns-2})  can be expressed as an averaging step within a
proximal step of the function $g$:
\begin{equation}
 {y}^{(k+1)} = \min_{\{{y}\}} 
g({y}) 
+ \frac {\rho R }{2}  \| \overline{x}^{(k+1)} + {y}  +  \overline{u}^{(k)} \| _{2}^{2}.  \label{eqn:admm-cns-4} 
\end{equation}

\subsection{\bfseries Accelerated Proximal Gradient}
\label{APG}

Proximal gradient (PG) and its accelerated version (APG)  \cite[Ch.~4]{parikh2014proximal}, first-order methods
with theoretical super-linear and quadratic convergence rates, have been widely applied to solve (\ref{eq:fun1}) due to their simplicity and adequate structure for solving large-scale problems. The standard PG method, which can be  easily  derived from a quadratic approximation model or a majorization-minimization approach  \cite{parikh2014proximal,beck2009gradient}, consists in iteratively computing the sequence
\begin{equation}
 {x}^{(k+1)}  =  prox_{\alpha g}\Big( {x}^{(k)} - \alpha_k \nabla
f({x}^{(k)}) \Big), \label{eqn:PGA-1}
\end{equation}
where  $\alpha_k \in [0, 1/L]$ denotes a suitable step size that is upperly bounded by the inverse Lipschitz constant. AGP  additionally includes  an extrapolation step (\ref{eqn:APG-2}) that produces a small correction of the gradient direction taking into account the information from the immediate past and current iteration. The algorithm is given by  
\begin{eqnarray}
 {x}^{(k+1)} &=&  prox_{\alpha g}\Big( {y}^{(k)} - \alpha \nabla
f({y}^{(k)}) \Big) \label{eqn:APG-1} \\
{y}^{(k+1)}  &=& {x}^{(k+1)} + \gamma_k (  {x}^{(k+1)} - {x}^{(k)})
 \label{eqn:APG-2}
\end{eqnarray}
 where $\gamma_k$, referred to as inertial sequence, is a weighting parameter that satisfies the following condition:
 \\
 
\begin{subequations}
\noindent
\hspace{-1.5mm}
\label{eqn:gamma-n-tk}
\begin{minipage}{0.5\textwidth}
\begin{equation}
\label{eqn:gamma_seq}
\gamma_k = \frac{t_{k} - 1}{t_{k+1}}, 
\end{equation}
\end{minipage}
\begin{minipage}{0.5\textwidth}
\vspace{-2mm}
\begin{equation}
\label{eqn:t_cond}
 t_{k+1}^2 - t_{k+1} \leq t_{k}^2 \text{\hspace{2.5mm}} \forall k \geq 1.
\end{equation}
\end{minipage}
\end{subequations}
\vspace{1.5mm}

\subsubsection{\bfseries Inertial sequence for the APG method} 
\label{APG:Inerseq}

Simple choices for the inertial sequence $\{ \gamma_k \}$, considering $t_1 = 1$, can be generated
using (\ref{eqn:various-iseq})\footnote{Other choices \cite{iutzeler-2018-proximal,liang-2018-faster} 
include ad-hoc rules or many more parameters.}:
Originally, \cite{Nesterov1983AMF} proposed to use 
(\ref{eqn:t-seq-Q}),
while more recently, among others, \cite{chambolle-2015-convergence,su-2016-differential,attouch-2016-rate} 
used (\ref{eqn:t-seq-lin}) for several values of $b \geq 2$ (being $b=2$ common practice). 
Furthermore, \cite{rodriguez-2019-improving} proposed a generalization of (\ref{eqn:t-seq-lin}), resulting 
in (\ref{eqn:t-seq-prop}), with $b=2$ and $a \in [50,\, 80]$ as default values.

\vspace{5mm}
\begin{subequations}
\noindent
\label{eqn:various-iseq}
\hspace{-2.5mm}
\begin{minipage}{0.5\textwidth}
\vspace{-2.5mm}
\begin{equation}
\label{eqn:t-seq-Q}
t_{k} = \frac{1 + \sqrt{1 + 4\cdot t_{k-1}^2} }{2},
\end{equation}
\end{minipage}
\begin{minipage}{0.5\textwidth}
\begin{equation}
\label{eqn:t-seq-lin}
t_{k} = \frac{k-1+b}{b}, \,~ b \geq 2,
\end{equation}
\end{minipage}
\begin{minipage}{0.495\textwidth}
\vspace{2.5mm}
\begin{equation}
\label{eqn:t-seq-prop}
t_{k} = \frac{k-1+a}{b}, \, ~b \geq 2,~ a \geq b-1. 
\end{equation}
\end{minipage}
\end{subequations}
\vspace{5mm}

\subsubsection{\bfseries Step-size for proximal gradient methods}  
\label{APG:step-size}

\textit{Exact/inexact line search}:
The exact line search defines $ \alpha_k = argmin_{\alpha} f( {x}_k - \alpha {g}_k )$, whereas for the inexact case $\alpha_k$ 
can be computed by some line search conditions, 
such as Goldstein, Wolfe or Armijo conditions \cite{nocedal-2006-numerical}.

\hspace{5mm} \textit{Barzilai-Borwein method}:
\label{sec:ss-BB}
\cite{barzilai-1988-two} proposed to use the information in the previous
iteration to estimate $\alpha_k$. Considering ${z}_k = {x}_k - {x}_{k-1}$  and ${r}_k =  \nabla f({x}_k) - \nabla f({x}_{k-1})$, \cite{barzilai-1988-two} proposed two variants, henceforth labeled BB-v1 (\ref{eqn:bb-v1}) and BB-v2 (\ref{eqn:bb-v2}), where $\langle \cdot , \cdot \rangle$ represents inner product, which can be shown to exhibit R-superlinear convergence for the Gradient method.
Unfortunately, for non-convex objective functions, these step-sizes
can result in a negative values. A positive step-size labeled BB-v3 (\ref{eqn:bb-v3}), which is exactly the geometric mean of the BB-v1 and BB-v2, was inferred using different methodologies \cite{vrahatis2000class,dai2006new,huang2017new}, being the simplest and earliest \cite{vrahatis2000class} obtained from the Lipschitz conditions. 

\vspace{5mm}
\begin{subequations}
\noindent
\begin{minipage}{0.33\textwidth}
\begin{equation}
\label{eqn:bb-v1}
\alpha_k = \frac{\langle {z}_k, {r}_k \rangle}{\| {r}_k \|_2^2},
\end{equation}
\end{minipage}
\begin{minipage}{0.33\textwidth}
\begin{equation}
\label{eqn:bb-v2}
\alpha_k = \frac{\| {z}_k \|_2^2}{\langle {z}_k, {r}_k \rangle},
\end{equation}
\end{minipage}
\hfill
\begin{minipage}{0.33\textwidth}
\begin{equation}
\label{eqn:bb-v3}
\alpha_k^2 = \frac{||z_k||_2^2}{||r_k||_2^2}. 
\end{equation}
\end{minipage}
\end{subequations}
\vspace{2.5mm}

\hspace{5mm}\textit{Cauchy step and variants:}
While it is well-known that the standard Cauchy step (\ref{eqn:cauchy-std}) can be inefficient
 and that it is always too long \cite[Sect. 3]{yuan-2008-stepsizes},
 there are successful variants: (i) in the context of sparse representations
\cite{blumensath-2008-iterative} proposed to use (\ref{eqn:cauchy-wSupp}), where
${s}_k = I_{[ |{x}_k  | > 0 ]}$, $I_{[{\tiny\mbox{COND}}]}$ represents the 
Indicator function\footnote{Equal to 1 if ``COND'' is true, 0 otherwise} and $\odot$ represents
element-wise product, (ii) in the context of convex quadratic optimization, \cite{dai-2006-new}
proposed (\ref{eqn:cauchy-mod}) and proved\footnote{\cite{dai-2006-new} also noticed that BB-v2 
or (\ref{eqn:bb-v2}) is the Cauchy step evaluated at the previous iteration $k$-$1$.}
that it asymptotically converges to (\ref{eqn:cauchy-std}).

\vspace{5mm}
\begin{subequations}
\noindent
\begin{minipage}{0.33\textwidth}
\begin{equation}
 \alpha_k = \frac{\| {g}_k \|_2^2}{\| \Phi {g}_k \|_2^2}, 
 \label{eqn:cauchy-std}
\end{equation}
\end{minipage}
\begin{minipage}{0.33\textwidth}
\begin{equation}
\alpha_k = \frac{\| {s}_k\odot{g}_k \|_2^2}{\| \Phi ({s}_k\odot{g}_k) \|_2^2}, 
\label{eqn:cauchy-wSupp}
\end{equation}
\end{minipage}
\begin{minipage}{0.33\textwidth}
\begin{equation}
\alpha_k^2 = \frac{\| {g}_k \|_2^2}{\| \Phi^T \Phi {g}_k \|_2^2}. 
\label{eqn:cauchy-mod}
\end{equation}
\end{minipage}
\end{subequations}

\subsection{\bfseries FISTA-3K : an improved FISTA variant}
\label{Fista-k3}
For a $\ell_1$-based regularization problem, i.e. $g(x) = \lambda ||x||_1$ in (\ref{eq:fun1}), one of the most convenient PG algorithms employed to perform the minimization would be Fast iterative shrinkage thresholding algorithm (FISTA) \cite{beck2009fast}. While its theoretical rate of convergence (RoC) is proportional to  $1/(\alpha t_k^2)$ \cite[Sect.~5.2]{becker2011templates}, due to some construction rules, the reported FISTA's RoC is equal to $\mathcal{O}(k^{-2})$. However, \cite{silva2019fista} proved that a bounded and non-decreasing step-size sequence $\alpha_k \leq \alpha_{k+1} $, can produce a cubic RoC in  small/medium intervals of indexes $k$, where the step-size approximately exhibits a linear growth. Moreover, the authors noticed that (\ref{eqn:cauchy-wSupp}) multiplied by small constant, denoted in (\ref{eq:Fk3}) by $c$, can generate the bounded and non-decreasing step-size sequence.
\begin{equation}
\alpha_k = c \frac{||s_k \odot g_k ||_2^2}{||\Phi (s_k \odot g_k)||_2^2} 
\label{eq:Fk3}
\end{equation}
\\ \vspace{-6mm}

%***************************************************************************************
%***************************************************************************************
%***************************************************************************************

\section{\bfseries Proposed method}
\label{sect:PGcons}

In this section, we generically describe and derive the theoretical support of our consensus model based on the proximal gradient method. %Additionally, we provide the details of how to estimate its correlative step-size in order to have a self-adjusting algorithm. 
\subsection{\bfseries Proximal Gradient Consensus}
\label{sect:PGcons2}

In (\ref{eq:cns1}), for convenience, we reproduced the objective function $F(x)$ presented in (\ref{eq:dist_reg}), which is composed by convex $L$-smooth functions $f_i$ and a convex function $g$.
\begin{equation}
\argmin _{\{ {x}\}}~
F(x) =
\fsum_{i=1}^{R} f_i({x}) +  g({x}). 
\label{eq:cns1}
\end{equation}
Minimization of each local function $f_i$,  which depends of the same global variable $x$, is not such a trivial case when looking for a descent solution. First, it should be posed in a consensus form:
\begin{equation} 
\argmin _{\{ {x}_i\}}\fsum_{i=1}^{R} f_i({x_i}) +  g({x_i})
\text{\hspace{5mm}s.t.\hspace{5mm}} ~ x_1 = x_2=\dots = x_R,
\label{eq:cns2}
\end{equation} 
where the global variable $x$ is replaced by local ones $x_i$  
using consistency constraint that enforces an agreement between variables. If we add a common auxiliary variable to the equality, an ADMM-consensus formulation can be directly obtained.
Alternatively, if we define a consensus set \cite[Ch.~5]{parikh2014proximal}  $C_c = \{({x}_{1}, {x}_{2}, \cdots, {x}_{R}) |   {x}_{1} = {x}_{2} = \cdots = {x}_{R}\}$
and using the indicator function of a set $S$
\begin{equation}
\iota_S(X) = \begin{cases}
0 &\text{if $X \in S$}\\
\infty &\text{if $X  \notin S$},
\end{cases}
\label{eq:indfunc}
\end{equation}
then the problem (\ref{eq:cns2}) can be written as a unconstrained consensus problem
\begin{equation} \argmin _{\{ {x}_i\}}~\fsum_{i=1}^{R} f_i({x_i}) +  g({x_i})
+ \iota_{C_C}({x}_{1}, {x}_{2},\cdots, {x}_{R}).
\label{eq:cns3}
\end{equation} 
Considering this latter formulation as starting point for an adequate minimization of the function $F(x)$, we use the quadratic approximation model \cite{beck2009gradient} on each differentiable function $f_i$
in order to derive the corresponding proximal gradient algorithm and its accelerated version:
 \begin{equation}
x^{(k+1)}_i = \argmin_{x_i}
 \frac{1}{2\alpha} \fsum_{i=1}^{R} \left\| {x}_i - \left(~{x}^{(k)}_i - \alpha  \nabla f_i({x}^{(k)}_i)~\right)\right\|_2^2 + g({x}_i) +  \iota_{C_C}({x}_{1}, {x}_{2}, \cdots, {x}_{R}).
\label{eq:cnsds} 
\end{equation}

\noindent Introducing a global auxiliary variable $y$ with
the constraint ${x_i - y = 0\text{\hspace{1.5mm}} \forall i}$,  the problem (\ref{eq:cnsds}) and its terms can be strategically rewritten as:
\begin{equation}
 x^{(k+1)}_i =  \argmin_{x_i, y}
  \frac{1}{2\alpha} \fsum_{i=1}^{R} || y - v_i^{(k)}||_2^2  + \frac{\rho}{2} \fsum_{i=1}^{R} ||y - x_i ||^2_2  + g(y)  + \iota_{C_C}({x}_{1},  {x}_{2}, \cdots, {x}_{R})  .
  \label{eq:cnsds-2var}  
\end{equation}
where  $v^{(k)}_i =  {x}^{(k)}_i - \alpha  \nabla f_i({x}^{(k)}_i)$. At first glance, the minimization summarized in (\ref{eq:cnsds-2var}) would  be an alternating optimization between two sub-problems 
\begin{eqnarray}
y^{(k+1)} &=& \argmin_{y}
 \frac{1}{2\alpha} \fsum_{i=1}^{R} || y - v^{(k)}_i||_2^2 
 + \frac{\rho}{2} \fsum_{i=1}^{R} ||y - x^{(k)}_i ||^2_2  + g(y)
\label{eq:cnsds-splitY}  \\
 x^{(k+1)}_i  &=& \argmin_{x_i}
 \frac{\rho}{2} \fsum_{i=1}^{R} ||y^{(k+1)} - x_i ||^2_2  + \iota_{C_C}({x}_{1},   \cdots, {x}_{R}),
\label{eq:cnsds-splitX}  
\end{eqnarray}
which involve selecting an additional penalty parameter and estimating two variables. Nevertheless, we show next that a simpler and effective solution is possible.
Via a direct algebraic manipulation, the \textit{ quadratic} terms of (\ref{eq:cnsds-splitY}) can be written as in (\ref{eq:cnsds-aux}) where $\epsilon$ represents terms that do not depend on $y$, and can be ignored when solving the optimization problem.
\begin{equation}
\frac{1}{2\alpha}  ||y||_2^2  - \frac{1}{\alpha} \langle y, v^{(k)}_i \rangle + \frac{\rho}{2}||y||_2^2  - \rho \langle y, x^{(k)}_i \rangle + \epsilon 
\label{eq:cnsds-aux}
\end{equation}
Discarding $\epsilon$, the remaining terms can be regrouped as follows
\begin{align}
 \frac{1 + \rho \alpha}{2\alpha}  ||y||_2^2  -  \langle y, \frac{1}{\alpha} v^{(k)}_i + \rho x^{(k)}_i \rangle   
 &= \frac{1 + \rho \alpha}{2\alpha} \left\|y  -  \frac{1}{1 + \rho \alpha} \left( v^{(k)}_i + \rho \alpha x^{(k)}_i\right)\right\|^2_2    \notag \\
& = \frac{1 + \rho \alpha}{2\alpha}  \left\|y  -  \frac{1}{1 + \rho \alpha}  \left( {x}^{(k)}_i - \alpha  \nabla f_i({x_i^{(k)}}) + \rho \alpha x^{(k)}_i\right)\right\|^2_2   \notag \\
& = \frac{1}{2\alpha_c}  \left\|y  -    \left(~ {x}^{(k)}_i - \alpha_c \nabla f_i({x_i^{(k)}}) ~ \right)\right\|^2_2  
\end{align}
where $\alpha_c = \alpha / (1 + \rho \alpha)$ is called consensus step-size. The new sub-problem, which is equivalent to (\ref{eq:cnsds-splitY}), is
\begin{eqnarray}
y^{(k+1)} = \argmin_{y}  
 \frac{1}{2\alpha_c}  \fsum_{i=1}^{R} \left\|y  -    \left(~ {x}^{(k)}_i -
  \alpha_c \nabla f_i({x_i^{(k)}})~ \right)\right\|^2_2  + g(y) .
\label{eq:cnsds2}
\end{eqnarray}
\noindent As the proximal mapping of (\ref{eq:cnsds2}) is of the form 
\begin{equation*}
prox_{\alpha g}\left(\frac{1}{R}\fsum_{j=1}^{R}{z}_j \right) = \argmin_{y} 
\frac {1}{2\alpha} \fsum_{j=1}^{R}  \left \lVert{  {y} -  {z}_j}\right \rVert _{2}^{2} +  g(y)
\end{equation*}
as defined in \cite{boyd2011distributed},
the $y$-update is given by
\begin{equation}
y^{(k+1)}  = prox_{\alpha_c g}\left(\frac{1}{R}\fsum_{i=1}^{R}\left(~{x}_i^{(k)} - \alpha_c  \nabla f_i({x}_i^{(k)})~\right)\right).
\label{eq:cns7}
\end{equation}

\noindent The $x$-update (\ref{eq:cnsds-splitX}) is just a projection onto the set $C_C$ 
\begin{equation}
x^{(k+1)}_i = P_{C_C}(Y^{(k+1)}) 
\end{equation}
where $Y^{(k+1)} = [y^{(k+1)},~y^{(k+1)},\dots,~y^{(k+1)}]^T$ and the projection operator 
$P_{C_C}(V) = [\overline{V},\overline{V}, \cdots,\overline{V}]^T$
averages all variables $v_i$ of $V$. Because $Y^{(k+1)}$ contains a single value
$y^{(k+1)}$, the projection calculation would be unnecessary, i.e. $x^{(k+1)}_i = y^{(k+1)}$. Therefore, the final solution is 
\begin{equation}
x_i^{(k+1)}  = prox_{\alpha_c g}\left(\frac{1}{R}\fsum_{i=1}^{R}\left(~{x}_i^{(k)} - \alpha_c  \nabla f_i({x}_i^{(k)})~\right)\right) ~.
\label{eq:cns7}
\end{equation}

Furthermore, we can obtain its fast version, termed APG consensus, by  adding the extrapolation step ${z}_i^{(k+1)}   = {x}_i^{(k+1)} + \gamma_k (  {x}_i^{(k+1)} - {x}_i^{(k)})$, and evaluating on (\ref{eq:cns7}) the variable $z_i$ instead of $x_i$.

Clearly, the presented consensus algorithm and its accelerated version do not
affect the rate of convergence, since  the initial assumptions (smoothness and 
convexity) of (\ref{eq:cns1}) and the resulting sequence (\ref{eq:cns7}), which can be compacted as
$ x_i^{(k+1)} = prox_{\alpha_c g}\left(~\overline{x}^{(k)} - \alpha_c  \overline{\nabla f}({x}_i^{(k)})~\right)$,
are identical to those of the standard  proximal gradient methods \cite[Ch.~10]{beck2017first}.

\subsection{\bfseries Step size for PG consensus based algorithms}
On what follows, for illustrative purposes, we only determinate the consensus step-sizes corresponding to the classic Cauchy and Barzilai-Borwein methods. 
Other step-sizes can be inferred from this explanation.  Considering the PG consensus algorithm  as $x_i^{(k+1)} = prox_{\alpha_c g}\left(\overline{x}^{(k)} - \alpha_c  \overline{g}^{(k)}\right)$, where $\overline{g}^{(k)} = \overline{\nabla f}({x}_i^{(k)})$ and an objective function $f_i(x_i^{(k)}) = \| \Phi_i x_i^{(k)} - b_i \|_2^2$, the Cauchy step-size, obtained from the exact line search, is formulated as
\begin{eqnarray}
 {\alpha_c}^{(k)}  = \argmin_{\alpha_c}
 \fsum_{i=1}^{R} || {\Phi}_i (\overline{x}^{(k)} - \alpha_c \overline{g}^{(k)}) -  {b}_i ||_2^2   
 %&\alpha \fsum_{i=1}^{R} || {A}_i  \overline{\nabla f}({x}) ||_2^2 - M \frac{ \fsum_{i=1}^{R} A_i^T(A_i \overline{x}  - b_i)}{M} \overline{\nabla f}({x}) = 0 \\
  = \frac{ \| \overline{g}^{(k)} \|^2_2 } { \Big(\frac{1}{R} \fsum_{i=1}^{R} \| {\Phi}_i  \overline{g}^{(k)} \|_2^2\Big)}.
 \end{eqnarray}
The Barzilai-Borwein step-sizes  are directly
computed from a two-point approximation on the secant equation (quasi-Newton strategy \cite{barzilai-1988-two}), defined as 
$\argmin_{D} ||\overline{z}^{(k)} - D \overline{r}^{(k)} ||_2^2 $ and
$\argmin_{D} ||D^{-1}\overline{z}^{(k)} -  \overline{r}^{(k)} ||_2^2,$
where $\overline{z}^{(k)} = \overline{x}^{(k)} - \overline{x}^{(k-1)}$  and $\overline{r}_k =  \overline{g}^{(k)} - \overline{g}^{(k-1)}$. From $D^{(k)} =  {\alpha_c^{(k)}} I$, the resulting BB-v1 and BB-v2 step-sizes are:
\vspace{5mm}

\begin{subequations}
\noindent
\begin{minipage}{0.49\textwidth}
\begin{equation}
{\alpha_c}^{(k)} = \frac{\langle \overline{z}^{(k)}, \overline{r}^{(k)} \rangle}{\| \overline{r}^{(k)} \|_2^2}, 
 \label{eqn:bb-v1-cs}
\end{equation}
\end{minipage}
\hfill
\begin{minipage}{0.49\textwidth}
\vspace{-2mm}
\begin{equation}
{\alpha_c}^{(k)} = \frac{\| \overline{z}^{(k)} \|_2^2}{\langle \overline{z}^{(k)}, \overline{r}^{(k)} \rangle}. \label{eqn:bb-v2-cs}
\end{equation}
\end{minipage}
\end{subequations}
\vspace{3mm}

%***************************************************************************************
%***************************************************************************************

\section{\bfseries Convolutional Dictionary learning}
\label{sect:CDL}
 {%\color{black} Convolutional sparse representation is an emerging model that encloses two optimizations problems: convolutional sparse coding (CSC) [ref], in which coefficient maps are estimated from a given dictionary, and convolutional dictionary learning (CDL) [ref], in which dictionary filters and coefficient maps are jointly estimated. 
The standard  formulation of CDL, extension of the Basis Pursuit DeNoising,  is posed as the optimization problem\footnote{Since the variables in (\ref{eqn:CDL-1}) correspond to 2D signals, the fidelity term would be intuitively calculated using a Frobenious norm. However, one can express these variables as 1D signals without losing generality and treat them with the $\ell_2$ norm, as it is done across the CSR literature.}: 
%the optimization problem \footnote{Since the variables in (1) correspond to 2D signals, the fidelity term would be intuitively calculated using a Frobenious norm. However, one can express these variables as 1D signals without losing generality and treat them with the $\ell_2$ norm, as it is done across the CSR literature.}:
\begin{equation}
\argmin _{ \{ {x}_{k,m}\} \{ {d}_{m}\} }  \frac{1}{2}  \fsum_{k} \norm{\fsum_{m}   {d}_m * {x}_{k,m} - {s}_k}_2^2 
+ \lambda \fsum_{k} \fsum_{m} \|{x}_{k,m} \|_1 {\hspace{5mm}s.t.\hspace{5mm}} \|{d}_m\|_2 \leq 1 \text{\hspace{2.5mm}} \forall m,
\label{eqn:CDL-1}
\end{equation}
\noindent where  $\{{x}_{k,m}\}$ represents set of coefficient maps, $\{{d}_m\}$ the set of dictionary filters, and $\{s_k\}$ is the set of training signals. The $\ell_2$ norm constraint of the filters is also required to avoid scaling ambiguities between filters and coefficient maps. This CDL problem (\ref{eqn:CDL-1}) has a non-convex  geometry when is evaluated in both variables $\{ {x}_{k,m}\}$ and $\{ {d}_{m}\}$. Nevertheless, by keeping either variable fixed, it can be recast as an alternating procedure of two convex problems:  coefficient update  and dictionary update.}

\subsection{\bfseries Coefficient update}

The coefficient update given by  (\ref{eq:CoefUp}) is a multiple measurement vector (MMV) \cite{eldar2015sampling} version of the CSC problem, where a collection of $K$ signals are simultaneously represented by the same dictionary.
\begin{equation}
\argmin _{ \{ {x}_{k,m}\}  }  \frac{1}{2}  \fsum_{k} \norm{\fsum_{m}   {d}_m * {x}_{k,m} - {s}_k}_2^2 
+ \lambda \fsum_{k}\fsum_{m} \|{x}_{k,m} \|_1
\label{eq:CoefUp}
\end{equation}
Defining  a Toeplitz matrix $ {D}_m$ such that $ {D}_m  {x}_{k,m} =  {d}_m *  {x}_{k,m}$, and two matrices
\begin{equation}  {D} = \left(\begin{array}{ccc}  {D}_0 &  {D}_1 & \ldots \end{array} \right) \text{\hspace{7.5mm}}  {X}_k = \left(\begin{array}{c} {x}_{k,0}\\  {x}_{k,1}\\ \vdots \end{array} \right), \end{equation} 
for convenience of notation, we can express (\ref{eq:CoefUp}) in a more simplified form 
\begin{equation}
\argmin _{ \{ {X}_{k}\}  }  \frac{1}{2}  \fsum_{k} \| {D} {X}_{k} - {s}_k \|_2^2 
+ \lambda \fsum_{k}\|{X}_{k} \|_1  
\label{eqn:CSC-2} 
\end{equation}

\subsection{\bfseries Dictionary update}

The dictionary update is the most computationally demanding sub-problem on CDL that can be interpreted as a convolutional form of Method of Optimal Directions (MOD) [31] with a normalization constraint: 
\begin{equation}
\argmin _{\{ {d}_{m}\} }  \frac{1}{2}  \fsum_{k} \norm{\fsum_{m}   {x}_{k,m}  * {d}_m  - {s}_k}_2^2 
\text{\hspace{5mm}s.t.\hspace{5mm}} \|{d}_m \|_2 \leq 1
\label{eq:DictUp}
\end{equation}

As we want to handle the convolutional component  of the fidelity term in the frequency domain, it is necessary to give an adequate spatial support to the target filters using zero-padding projection operator $P$. Mixing this spatial support and the normalization requirement, the constraint set is given by
\begin{equation}
C_{\text{PN}} = \{{z} \in {\rm I\!R}^N : (I - PP^T){z} =0, \left\| {z} \right\|_2 = 1 \} \;.
\end{equation} 
Employing the indicator function\footnote{Generic indicator function previously defined in Eq. (\ref{eq:indfunc}).} $\iota _{C_{\text {PN}}}(\cdot)$ of the constraint set $C_{\text{PN}}$, the dictionary update can be written in unconstrained form
\begin{equation}
\argmin_{\{ {d}_{m}\} }  \frac{1}{2}  \fsum_{k} \norm{\fsum_{m}   {x}_{k,m} * {d}_m  - {s}_k}_2^2  + \fsum_{m} \iota _{C_{\text {PN}}}( {d}_{m}) \;.
\label{eq:DictUp2}
 \end{equation}
If we define a Toeplitz matrix $ {X}_{k,m}$ such that $ {X}_{k,m}  {d}_m =   {x}_{k,m} *  {d}_m$  , and the matrices
\begin{equation} 
 {X}_k = \left(\begin{array}{ccc}  {X}_{k,0} &  {X}_{k,1} & \ldots \end{array} \right) \text{\hspace{7.5mm}}  {D} = \left(\begin{array}{c} {d}_{0}\\  {d}_{1}\\ \vdots \end{array} \right),    \end{equation} 
we can rewrite the unconstrained problem (\ref{eq:DictUp2}) as
\begin{equation}
\argmin _{ \{ {D}\}  }  \frac{1}{2}  \fsum_{k} \| {X}_{k} {D}  - {s}_k \|_2^2 
+  \iota_{C_{PN}}({D})  
\label{eq:DictUp3}
\end{equation}
or, by collecting
\begin{equation} 
 {X} = \left(\begin{array}{c} {X}_{0}\\  {X}_{1}\\ \vdots \end{array} \right) \text{\hspace{7.5mm}}  {S} = \left(\begin{array}{c} {s}_{0}\\  {s}_{1}\\ \vdots \end{array} \right)   \end{equation} as
\begin{equation}
\argmin _{ \{ {D}\}  }  \frac{1}{2}  \| {X}  {D} - {S} \|_2^2 
+  \iota_{C_{PN}}({D})  
\label{eq:DictUp4}
\end{equation}

While there are many strategies in which the CDL problem (\ref{eqn:CDL-1}) can be  addressed, all of them can be classified in two categories: way of dealing with convolutions (frequency domain and separable filters), and  way of processing data (batch, mini-batch and online methods). The algorithms studied in the following sections will be batch methods raised in the frequency domain.

% ************************************************************************************
% ************************************************************************************

% ************************************************************************************
% ************************************************************************************

\section{\bfseries Coefficient update algorithms}
\label{sect:CoefUp_alg}

The coefficient update can be interpreted as 
$K$  CSC problems since each set of coefficient maps $X_k$ (\ref{eqn:CSC-2}) can be estimated independently. For the standard CSC problem, the ADMM algorithm in the frequency domain strikes a good trade-off between convergence and run-time while the FISTA algorithm in frequency domain has the advantage of low complexity in terms of number of operations. Nevertheless, \cite{silva2019fista}, summarized in Section \ref{Fista-k3},  has proved the existence of a better FISTA's rate of convergence, allowing to achieve a more adequate balance between the mentioned features.

\subsection{\bfseries ADMM}
The problem (\ref{eqn:CSC-2}) can be solved via ADMM approach by adding an auxiliary variable ${Y}_k$ that is constrained to be equal to the primary one ${X}_k$, i.e.
\begin{equation} \argmin _{\{  {X}_k\},\{  {Y}_k\}}~\frac {1}{2}  \fsum_{k} \lVert { {D}  {X}_k -  {s}_k} \rVert _{2}^{2}  + \lambda   \fsum_{k} \left \|   {Y}_k \right  \|_{1}  \text{\hspace{2.5mm}s.t.}\text{\hspace{2.5mm}}~  {X}_k =  {Y}_k. 
\label{eq:ADMM-0}
\end{equation} 

The associated ADMM iterations, in which the sum notation of $K$ images is removed to analyze (\ref{eq:ADMM-0}) as independent steps, are given by
\begin{eqnarray}
  {X}_k^{(i+1)}&=&\argmin _{\{  {X}_k\}} \frac {1}{2}   \lVert {  {D}  {X}_k -  {s}_k} \rVert _{2}^{2} +\,\frac {\rho }{2}    \lVert{  {X_k} -  {Y_k}^{(i)} +  {U}_k^{(i)}} \rVert _{2}^{2} \label{eq:ADMM-1} \notag \\ \\
  {Y}_k^{(i+1)}&=&\argmin _{\{  {Y}_k\}} \text{\hspace{1.5mm}} \lambda   \lVert{  {Y}_k}\rVert _{1}+\,\frac {\rho }{2}  \lVert{  {X}_k^{(i+1)} -  {Y}_k +  {U}_k^{(i)}} \rVert _{2}^{2} \label{eq:ADMM-2}  \\
 {U}_k^{(i+1)}&=&{U}_k^{(i)} +  {X}_k^{(i+1)} -  {Y}_k^{(i+1)}.  \label{eq:ADMM-3}
\end{eqnarray}

 The Y-update  (\ref{eq:ADMM-2}) has a closed form solution defined using soft-thresholding function
\begin{equation} 
 {Y}^{(i+1)} = T_{\lambda /\rho }\left ({  {X}^{(i+1)} +  {U}^{(i)} }\right ) \!, 
\label{eq:8}
\end{equation} 
where  $ T_{\gamma }(x) = sign(x) \odot max(0, |x| - \gamma)$. 

 In the spatial domain, the X-update (\ref{eq:ADMM-1}) implies the inversion of a linear system with high computationally expenses.  However, (\ref{eq:ADMM-1}) can be transformed into the frequency domain using means of the convolution theorem in order to get a new system with less complexity 
\begin{equation}
( {\hat {D}}^{H} { \hat {D}} + \rho I) \hat {  {X}}_k = {\hat {D}}^{H} \hat {  {s}}_k + \rho (\hat {  {Y}}_k -  \hat {  {U}}_k), 
\label{eq:10}
\end{equation}
\noindent where $\hat {  {s}}$ , $ \hat {  {D}}$,  $\hat {  {Y}}$ and  $\hat {  {U}}$ 
denote the frequency domain variables that are obtained after applying the DFT operator to the variables  $  {s}$ , $ {  {D}}$,  ${  {Y}}$ and  ${  {U}}$.  The matrix $ {\hat {D}}$ has a block structure of $M$ concatenated $N \times M$ diagonal matrices.

As the operation  $ {\hat {D}}^{H} { \hat {D}}$  results in a large matrix (of size  $MN \times MN$) with many zero values that are  not  part  of  the  final solution, \cite{wohlberg2014efficient} noted that it is only necessary to solve $N$ independent linear system of $M \times M$. Each independent system consists of a single sum between a rank-one term and diagonal term, which inversion can easily be  performed by applying the Sherman-Morrison formula on its rearranged form of non-zero elements. 

\subsection{\bfseries FISTA}

The standard FISTA algorithm for solving (\ref{eqn:CSC-2}) can be expressed  as
\begin{eqnarray}
{X}_k^{(i+1)} &=&  T_{\alpha \lambda  }\left(\Big [{X}_k + \alpha_k \nabla F\Big(\frac{1}{2}  \| {D}  {X}_k - {s}_k \|_2^2 \Big)\Big]_{X=Y^{(i)}}\right) \\
t^{(i+1)} &=& \frac{1}{2}\left(1+\sqrt{1 + 4(t^{(i)})^2}\right) \\
{{Y}}_k^{(i+1)} &=& {{X}}_k^{(i+1)} + \frac{t^{(i)}-1}{t^{(i+1)}}({{X}}_k^{(i+1)} - {{X}}_k^{(i)}) ,
\end{eqnarray}

\noindent where the gradient calculation of fidelity term $\frac{1}{2}  \| {D}  {X}_k - {s}_k \|_2^2$ can be computationally demanding due to the convolution operations. Nevertheless, \cite{wohlberg2015efficient} proposed to efficiently compute the gradient in the frequency domain as
\begin{eqnarray}
\nabla F \left( \frac{1}{2} \|{  {\hat D}  {\hat X}_k - {\hat s}_k}\|_2^2\right) =  {\hat D}^H({\hat D}  {\hat X}_k - {\hat S}).  
\label{eq:11}
\end{eqnarray}
\vspace{3.5mm}
\vspace{-5mm}

% ************************************************************************************
% ************************************************************************************
% ************************************************************************************
% ************************************************************************************

\section{\bfseries Dictionary update algorithms}
\label{sec:Dictup}
The dictionary update could be extremely expensive with respect to the coefficient update when the training set is large. Particularly, the first efficient methods, commonly based on ADMM, exhibited a computational complexity proportional to  $\mathcal{O}(K^2)$, where $K$ is the number of training images. The most recent ones, presented below, stood out for having a linear complexity.

\subsection{\bfseries ADMM consensus}
 As proposed in  \cite{vsorel2016fast}, the dictionary update problem (\ref{eq:DictUp3}) can be expressed in ADMM consensus form as
\begin{eqnarray} 
\mathop{\arg\min}\limits_{ {D}_k} \ \frac{1}{2}  \sum _k \Vert { {X}_k  {D}_k -  {s}_k}\Vert _2^2  +   \iota _{C_{\text{PN}}} ( {G}) \text{\hspace{2.5mm}s.t.\hspace{2.5mm}}    {D}_{k} =  {G}  \text{\hspace{2mm}}  \forall k\
\label{eq:ADMMcns}
\end{eqnarray} 
where \{$ {D}_{k}$\} is the local dictionary  for each training image and \{$ {G}$\} is the global consensus variable. The associated ADMM updates are
\begin{eqnarray}
  {D}_k^{(i+1)} &=& \argmin _{\{  {D}_k\}} \frac {1}{2}   \lVert {  {X}_k  {D}_k -  {s}_k} \rVert _{2}^{2} +\,\frac {\sigma }{2}    \lVert{  {D}_k -  {G}^{(i)} +  {H}_k^{(i)}} \rVert _{2}^{2}  \notag \\ \label{eq:ADMMcns-1} \\
  {G}^{(i+1)} &=& \argmin _{\{  {G}\}} \text{\hspace{1.5mm}}  \iota_{C_{PN}}({G}) +\,\frac {\sigma }{2} \fsum_{k}  \lVert{  {D}_k^{(i+1)} -  {G} +  {H}_k^{(i)}} \rVert _{2}^{2}  \notag\\\label{eq:ADMMcns-2} \\
 {H}_k^{(i+1)} &=&  {H}_k^{(i)} +  {D}_k^{(i+1)} -  {G}^{(i+1)}. \label{eq:ADMMcns-3}
\end{eqnarray}

The update (\ref{eq:ADMMcns-1}) can be decomposed in $K$ independent linear systems that are efficiently solved via the DFT domain Sherman Morrison method. The update (\ref{eq:ADMMcns-2})  has a closed-form solution 
\begin{eqnarray} 
 {G}^{(i+1)}  = {prox}_{\iota _{C_{\text {PN}}}} \Big (\frac{1}{K} \sum_k (  {D}_k^{(i+1)} +   {H}_k^{(i)}) \Big) \text{\hspace{1.5mm}.}
\label{eq:22}
\end{eqnarray}

Additionally, \cite{garcia-2017-subproblem}  introduced a more complete ADMM consensus algorithm to improve practical convergence by finding the best coupling variables ($X_k, Y_k, D_k$ or $G$) that are passed between the coefficient update and the dictionary update. Building on this, \cite{garcia2018convolutional} developed a multi-core implementation in order to fully exploit the independent and separable structure of the consensus algorithm.

\subsection{\bfseries APG}
\cite{garcia2018convolutional,silva2018efficient} proposed frequency domain based APG algorithms for solving (\ref{eq:DictUp4}), which basic scheme is 
\begin{eqnarray}
{D}^{(i+1)} &=&  {prox}_{\lambda}\left(\Big [{D} + \alpha \nabla F\Big(\frac{1}{2}  \| {X}  {D} - {S} \|_2^2 \Big)\Big]_{D=G^{(i)}}\right) \\ 
t^{(i+1)} &=& \frac{1}{2}\left(1+\sqrt{1 + 4(t^{(i)})^2}\right) \\
{{G}}^{(i+1)} &=& {{D}}^{(i+1)} + \frac{t^{(i)}-1}{t^{(i+1)}}({{D}}^{(i+1)} - {{D}}^{(i)})
\end{eqnarray}

\noindent \text{\hspace{-0.5mm}}In both algorithms, the gradient was computed in the frequency domain as
\begin{eqnarray}
\nabla F \left( \frac{1}{2} \|{  {\hat X}  {\hat D} - {\hat S}}\|_2^2\right) =  {\hat X}^H({\hat X}  {\hat D} - {\hat S}) \text{\hspace{1.5mm},}  
\label{eq:11}
\end{eqnarray}
 In order to avoid a grid search for the step-size selection, \cite{silva2018efficient} introduced an adaptive step-size estimation with low computational expenses given by 
 \begin{equation}
 \alpha  = \argmin_{\alpha}
|| {\hat X} ({\hat D} - \alpha {\hat g}) -  {\hat S} ||_2^2 = 
   \frac{||{\hat g}||^2_2}{ || {\hat X}  {\hat g} ||_2^2},
   \label{step-aut}
 \end{equation}
where ${\hat g} = \nabla F ( \frac{1}{2} ||  {\hat X}  {\hat D} - {\hat S}||_2^2)$. In comparison to the ADMM consensus algorithm, APG algorithm has a simpler structure; however, it does not have an intrinsic separability, which allows to compute components independently.

\subsection{\bfseries Proposed APG consensus}

As our previous work \cite{silva2018efficient_icip}, the dictionary update problem (\ref{eq:DictUp3}) can be posed in the following consensus form
\begin{equation} 
\argmin _{\{  {D}_{k}\}} \frac {1}{2}  \sum _{k}  \lVert  
 {X}_{k}   {D}_{k}  -   {s}_{k} \rVert _{2}^{2} 
+  \sum _{k} \iota _{C_{\text {PN}}} (  {D}_{k}) 
+\iota _{C_{\text {C}}} (  {D}_{1},  {D}_{2}, \cdots,  {D}_{K}) \text{\hspace{1.5mm}.}
\end{equation}
where $C_c = \{({D}_{1}, {D}_{2}, \cdots, {D}_{K}) |   {D}_{1} = {D}_{2} = \cdots = {D}_{K}\}$
is a  constraint set used to induce equality between local dictionaries. The corresponding APG consensus iterations, theoretically justified in Section \ref{sect:PGcons2}, are given by
\begin{eqnarray}
{D}_k^{(i+1)} &=&  \Big [{D}_k + \alpha_c \nabla F_k\Big(\frac{1}{2}  \| {X}_k  {D}_k - {s}_k \|_2^2 \Big)\Big]_{D=G^{(i)}} \\ 
{H}^{(i+1)} &=& {prox}_{\iota _{C_{\text {PN}}}} \left (\frac{1}{K} \sum_k  {D}^{(i+1)}_k \right) \\
t^{(i+1)} &=& \frac{1}{2}\left(1+\sqrt{1 + 4(t^{(i)})^2}\right) \\
{{G}}^{(i+1)} &=& {{H}}^{(i+1)} + \frac{t^{(i)}-1}{t^{(i+1)}}({{H}}^{(i+1)} - {{H}}^{(i)})
\end{eqnarray}
Analogous to the previous APG algorithm, we efficiently calculate each local gradient, denoted in (\ref{eq:11}), and  most of the algorithm components\footnote{See \cite{silva2018efficient_icip} for full implementation details.} in the frecuency domain. 
\begin{equation}
\nabla F_k \left( \frac{1}{2} \|{  {\hat X}_k  {\hat D}_k - {\hat s}_k}\|_2^2\right) =  {\hat X}_k^H({\hat X}_k  {\hat D}_k - {\hat s}_k) \text{\hspace{1.5mm}}  
\label{eq:11}
\end{equation}
\vspace{2.5mm}
\vspace{-5mm}

\section{\bfseries Computational Results}
\label{Sect:results}
Our experiments were carried out on
a desktop computer equipped with an Intel i7-7700K CPU (4.20 GHz, 8MB Cache, 32GB RAM) and a Nvidia Tesla P100 GPU card.

\subsection{\bfseries Results in convolutional dictionary learning task}
\label{Result-CDL}
In this Section \ref{Result-CDL}, we assess
in terms of convergence and computational performance the following CDL implementations: 
\begin{itemize}
\item \textbf{ADMM-ADMMCns}:  The CDL algorithm presented in \cite{garcia-2017-subproblem}, where ADMM  and ADMM consensus methods were used to solve the coefficient and dictionary updates respectively.
\item \textbf{P-ADMM-ADMMCns}: Parallel implementation of the ADMM-ADMMCns algorithm.
\item \textbf{FISTA-APGCns}: We proposed a CDL algorithm that consists of a FISTA-3K method and our APG consensus method  for solving the coefficient update and dictionary update respectively. 
\item \textbf{P-FISTA-APGCns}: Parallel implementation of the
FISTA-APGCns algorithm.
\end{itemize}
Both regular ADMM-ADMMCns and FISTA-APGCns algorithms are vectorized implementations coded in MATLAB while their parallel implementations are CUDA-enabled MATLAB codes\footnote{MATLAB implementations that exploit the convenience of the gpu-
Arrays and high-level GPU operations.}. The initial penalty parameters $\rho$ and $\sigma$ of ADMM and ADMM-consensus are selected from the heuristic rule presented in \cite{wohlberg2015efficient}. In our implementations, the FISTA-3K method uses the weighted Cauchy-support step-size, denoted in (\ref{eq:Fk3}), with a conservative multiplicative factor $c=0.2$ and the APG-consensus method uses a consensus step-size based on BB-v3.
To provide a fair comparison with our algorithms, which have adaptive parameters (automatic estimation of step-sizes), the ADMM algorithms use an update rule for the penalty parameters and an over-relaxation strategy, as described in Section \ref{Ref-ADMM-par}. 

In this first experiment, as training set, we used $K = \{5, 10, 20, 40\}$  gray-scale images of size $256\times 256$ pixels, cropped and rescaled from a set of images obtained from the MIRFFLICKR-IM dataset \cite{huiskes2010new}. For each CDL algorithm, a dictionary of $M$ filters of size $8 \times 8$ was learning using a sparsity parameter $\lambda = 0.1$ and a fixed number (1000) of iterations.

\begin{figure}[H]
\centering
\includegraphics[width=0.29\textwidth]{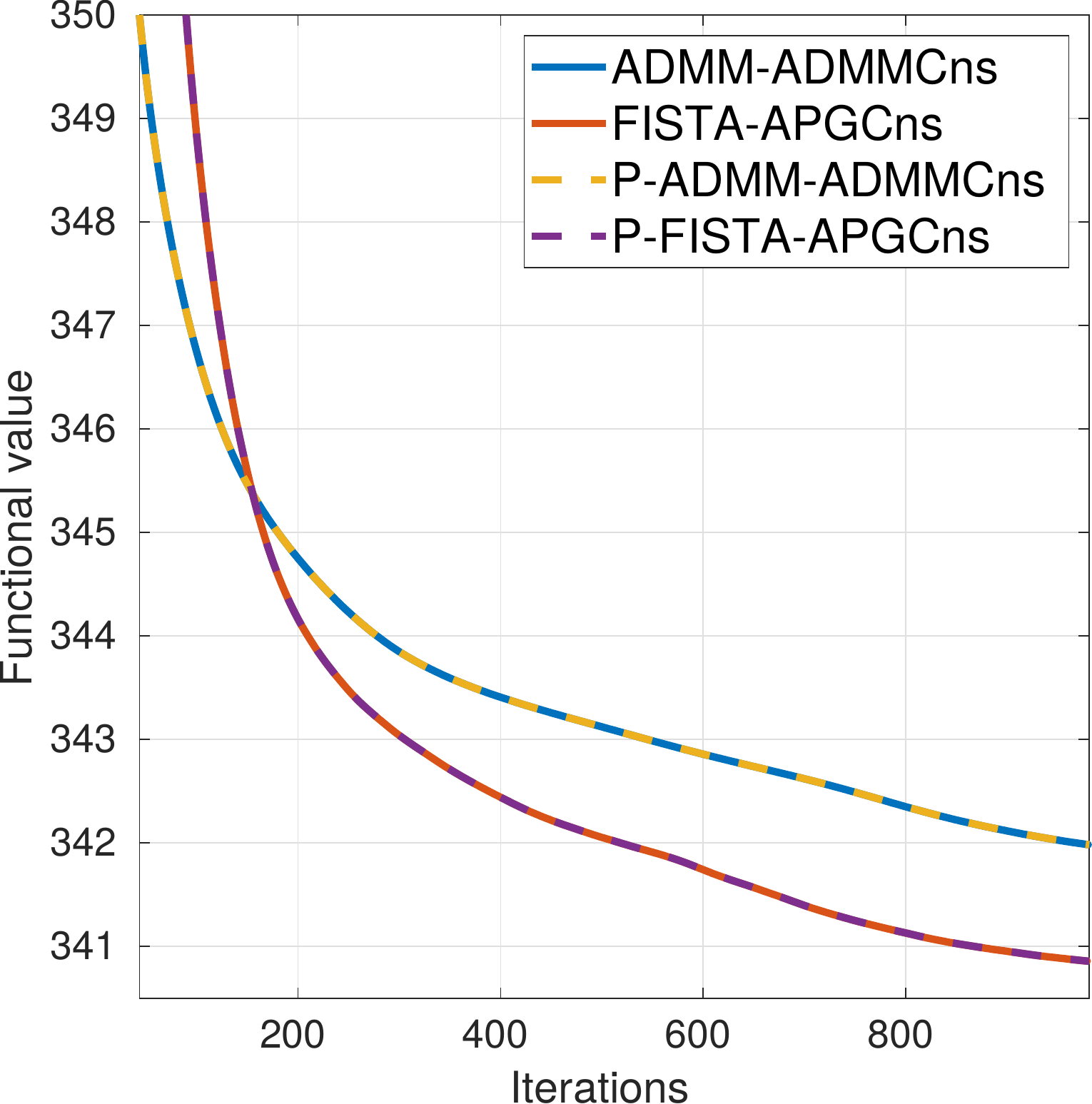}  
\hspace{5mm}
\includegraphics[width=0.28\textwidth]{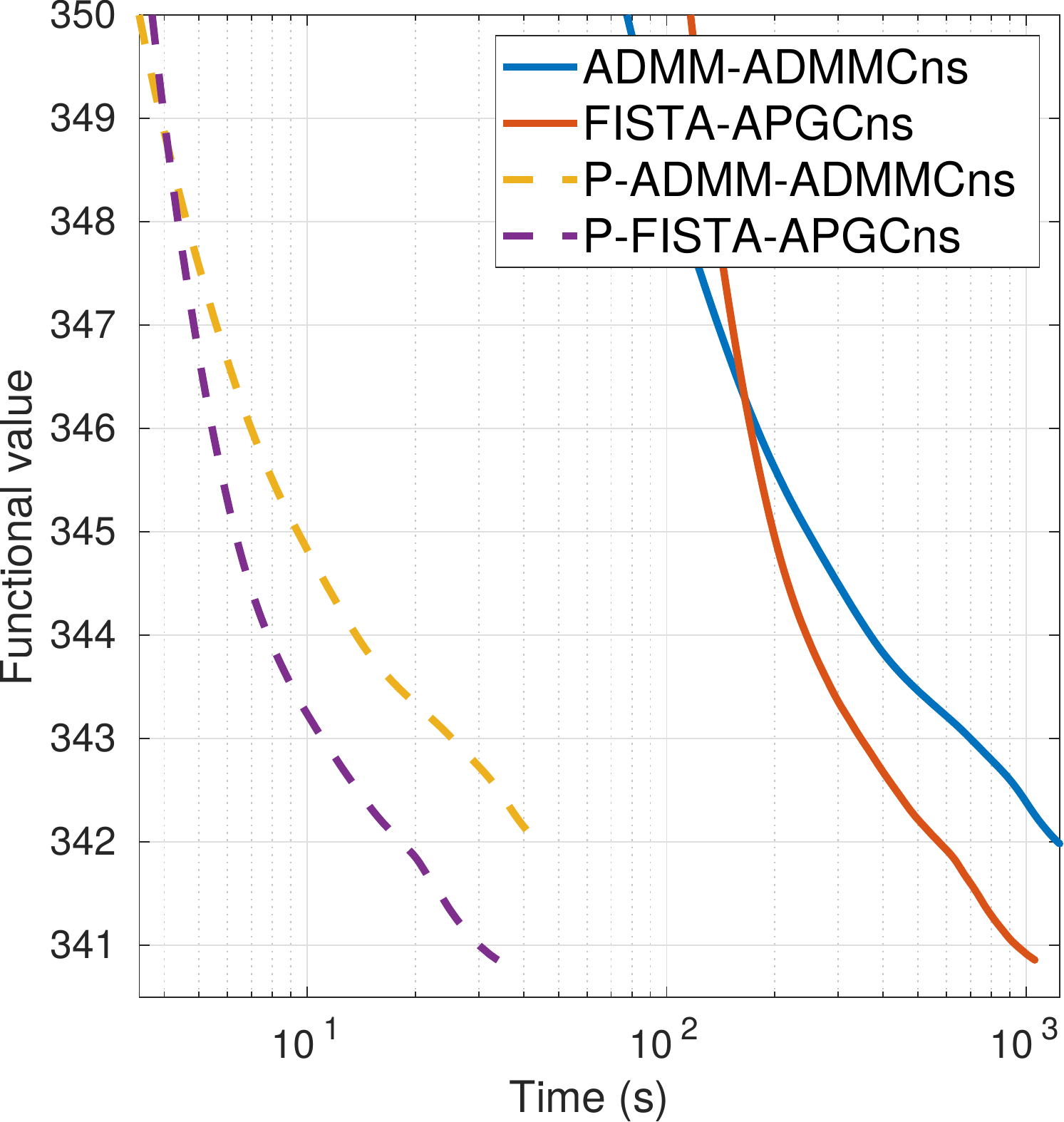} 
\caption[]{Convolutional dictionary learning: A comparison on a set of 5 training images of the functional value decay with respect to number of iterations and run-time\footnotemark.}%
\label{fig:cost_iter_img1}
\end{figure}

\begin{figure}[H]
\centering
\includegraphics[width=0.29\textwidth]{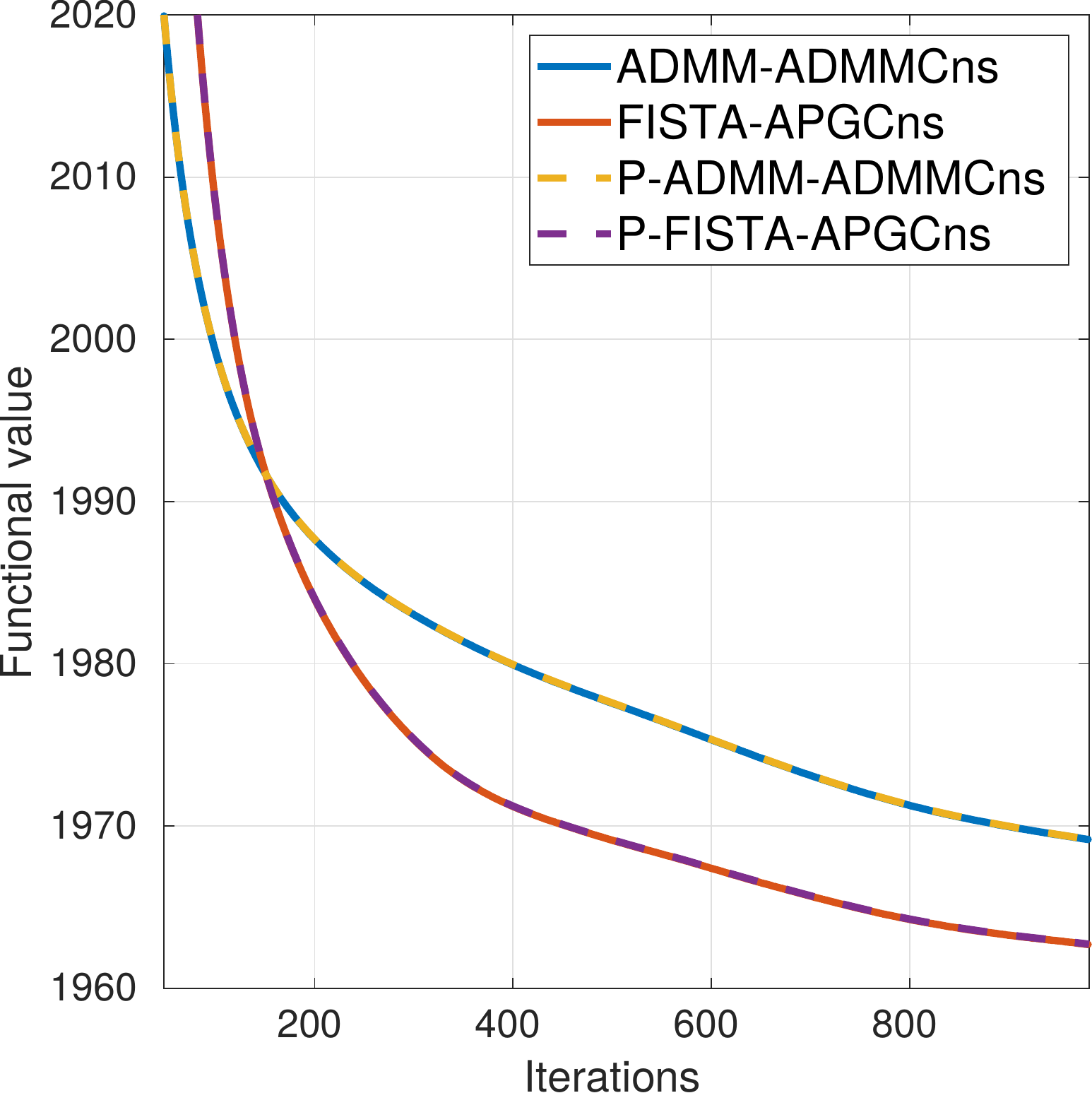}  
\hspace{5mm}
\includegraphics[width=0.28\textwidth]{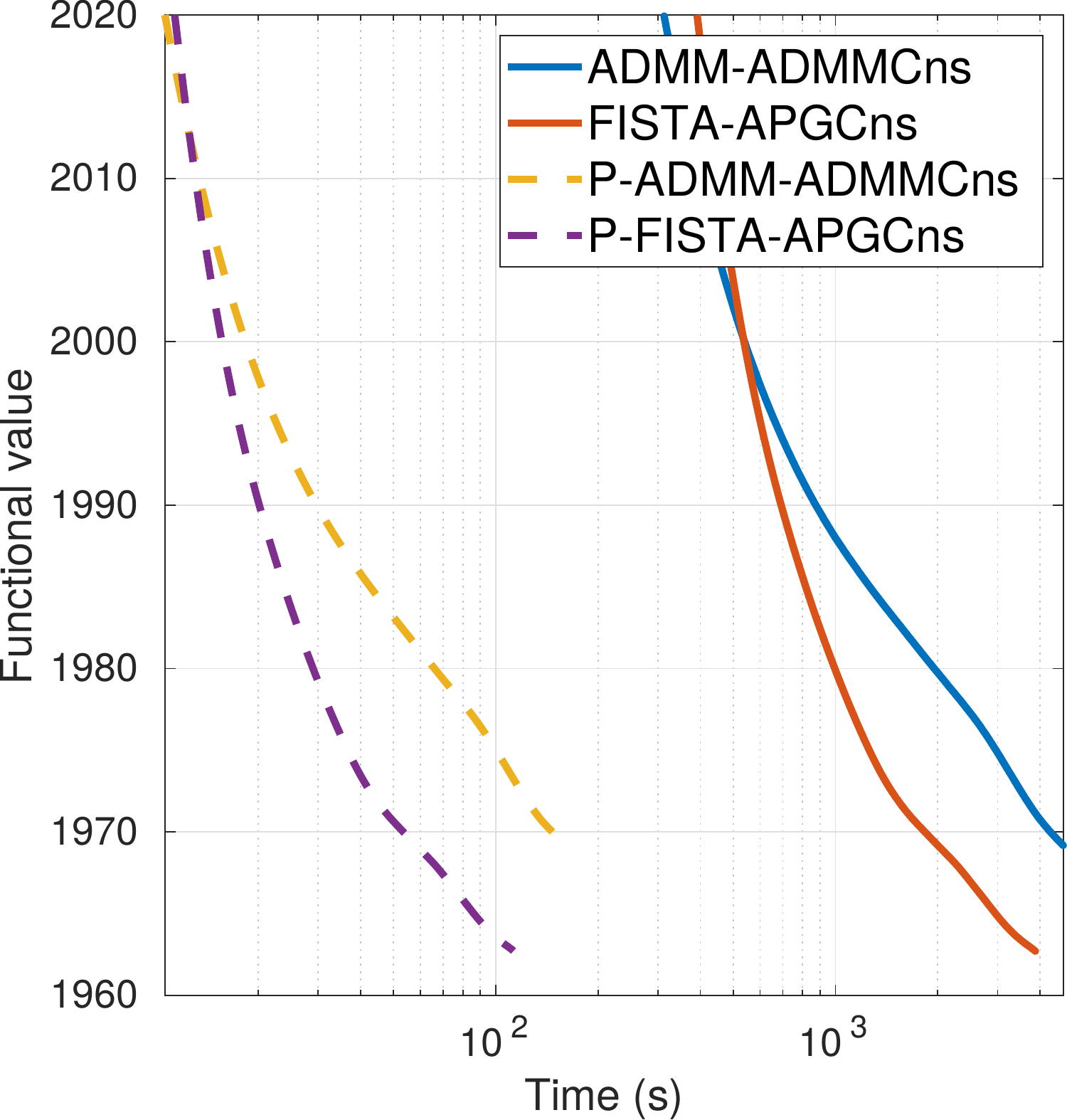} 
\vspace{-1mm}
\caption{Convolutional dictionary learning: A comparison on a set of 20 training images of the functional value decay with respect to number of iterations and run-time$^\text{\ref{note1}}$.}
\label{fig:cost_iter_img2}
\end{figure}

\begin{figure}[H]
\centering
\includegraphics[width=0.29\textwidth]{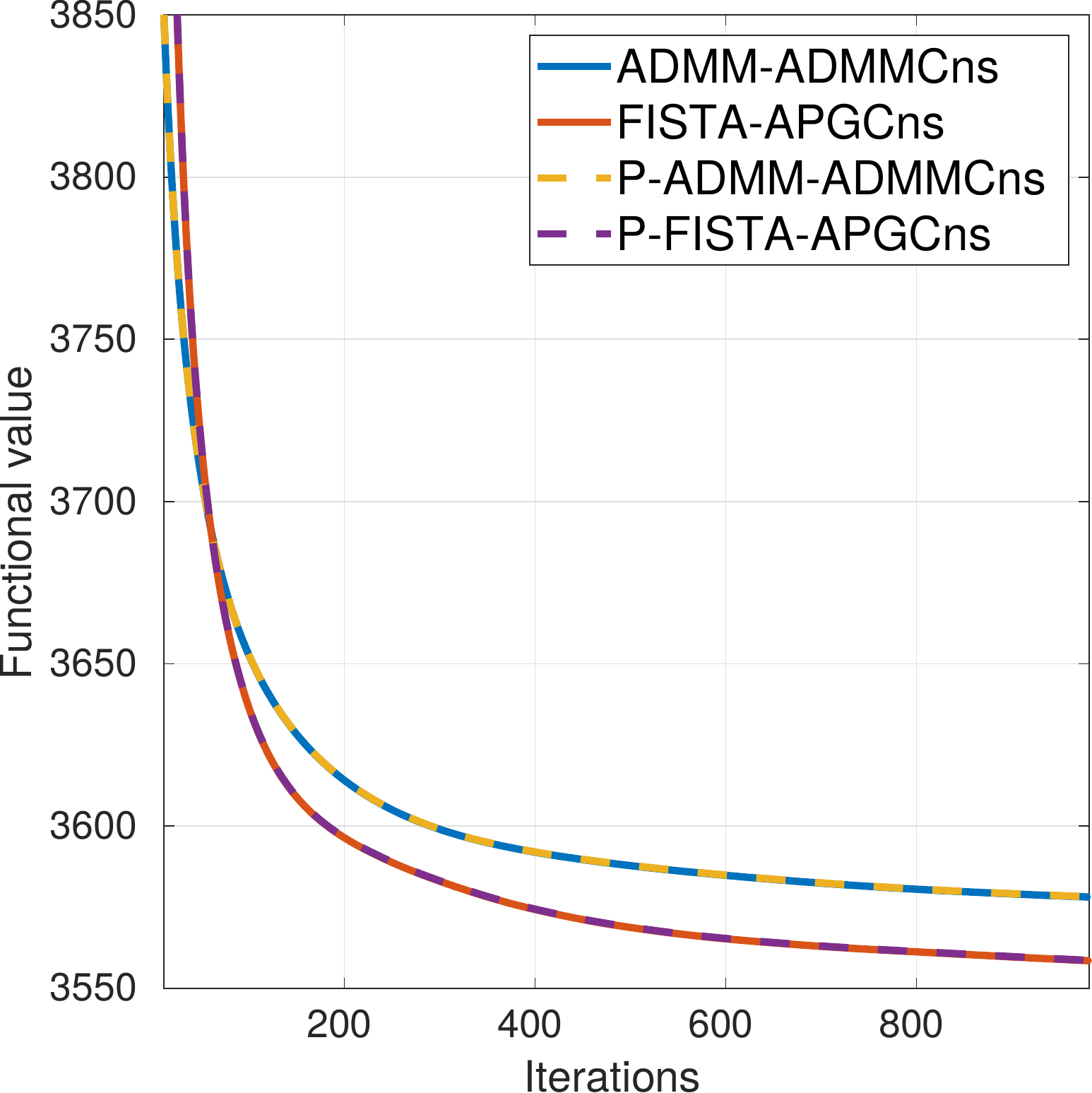}  
\hspace{5mm}
\includegraphics[width=0.28\textwidth]{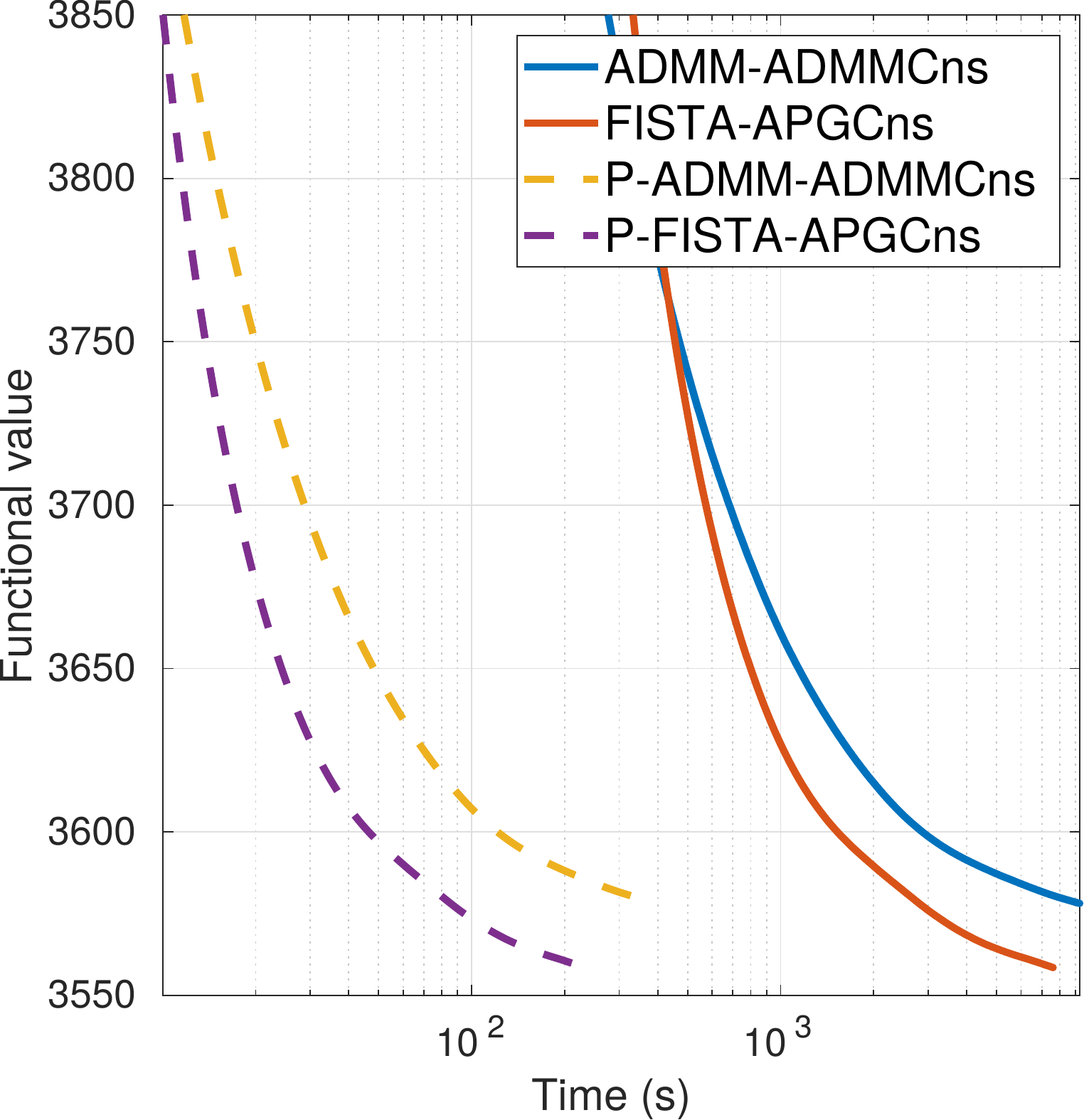} 
\vspace{-1mm}
\caption{Convolutional dictionary learning: A comparison on a set of 40 training images of the functional value decay with respect to  number of iterations and run-time$^\text{\ref{note1}}$.}
\label{fig:cost_iter_img3}
\end{figure}

\begin{figure}[H]
 \centering
 \subfigure[36 fixed dictionary filters]{\includegraphics[width=0.29\textwidth]{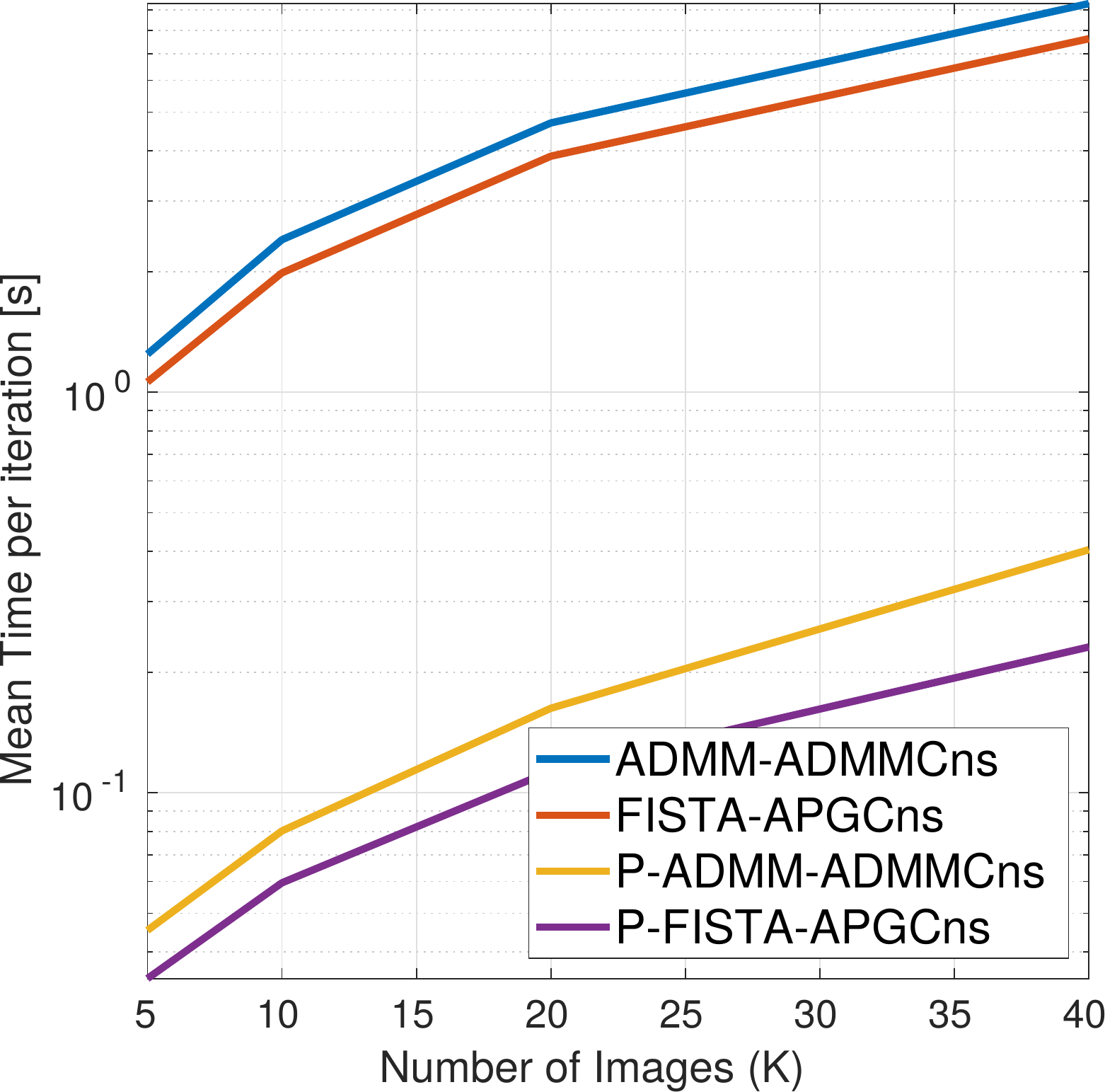} \label{subfig:fig3}} 
 \hspace{5mm}
 \subfigure[20 fixed training images]{\includegraphics[width=0.29\textwidth]{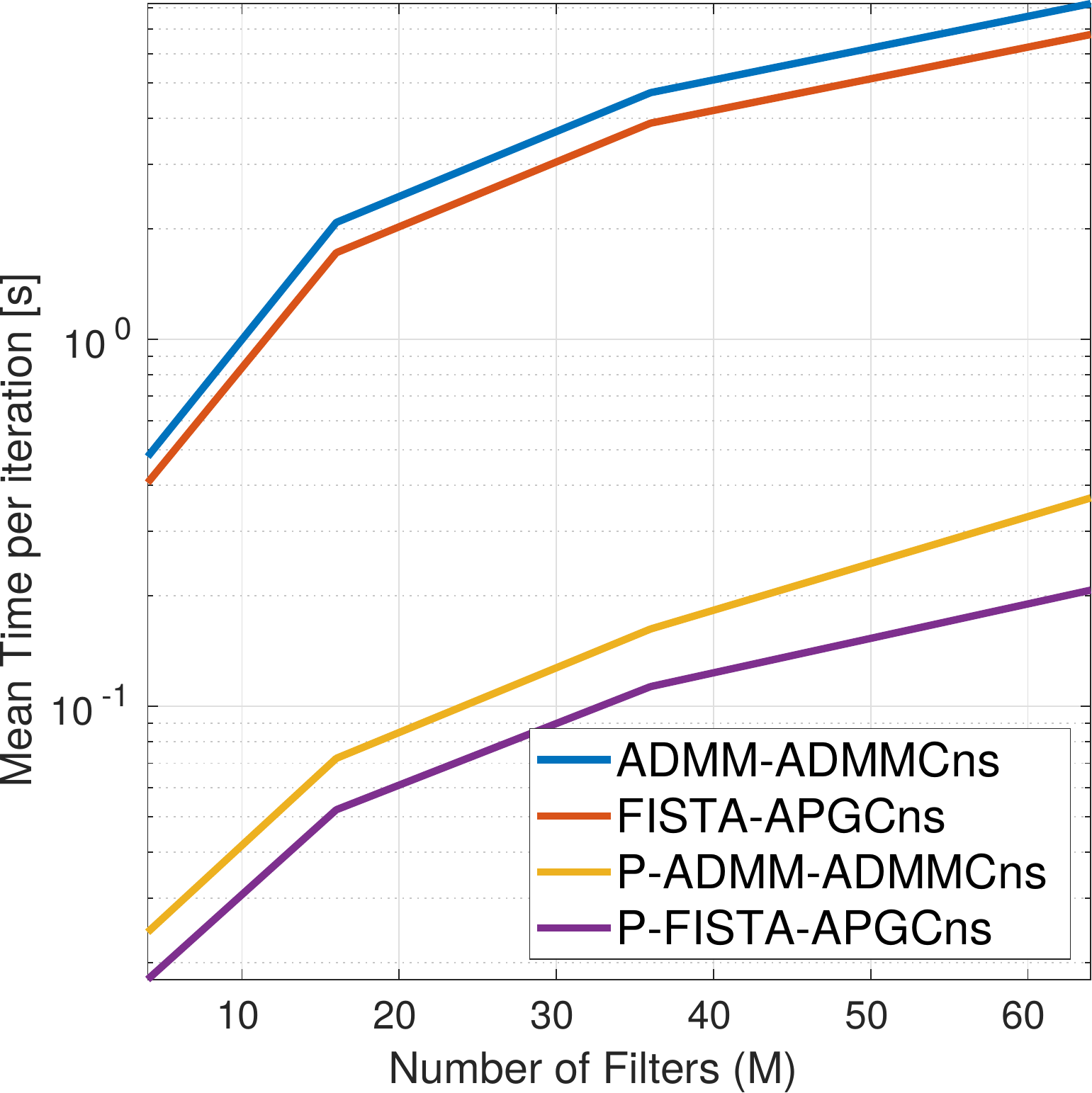} \label{subfig:fig4}}
  \vspace{-2mm}
 \caption{Convolutional dictionary learning: A comparison of mean time$^\text{\ref{note1}}$ per iteration when varying (a) training set size or (b) filter set size, and keeping the other one constant.}
 \label{fig:mean_time_filter_img}
 \end{figure}

In Figures \ref{fig:cost_iter_img1} to \ref{fig:cost_iter_img3}, we compare the performance of the implementations in learning a
dictionary of $36$ filters from different sizes of training sets.
Processing time of each implementation is summarized in Figure \ref{fig:mean_time_filter_img} and fully detailed in Tables \ref{tab:mean_time_img} and \ref{tab:mean_time_filter}.

\vspace{-4mm}
\vspace{-4mm}
\footnotetext{\label{note1}Time axis plotted in logarithmic scale.}

As can be observed in the Figures \ref{fig:cost_iter_img1} to \ref{fig:cost_iter_img3}, for a short range of iterations, the ADMM-ADMMCns algorithms initially have a better functional value decay. However, after that interval, our algorithms consistently outperform the ADMM-ADMMCns algorithms in terms of convergence with respect to both iterations and run-time.

\begin{table}[H]
\centering
\caption{Average time (in seconds) per iteration of each stage of the CDL algorithms varying the training set size ($K$).}
\scalebox{0.9}{
\renewcommand{\arraystretch}{1.15}
\addtolength{\tabcolsep}{-3pt}
\begin{tabular}{c|c|c|c|c|c|c|c|c|}
\cline{2-9}
                                           & \multicolumn{2}{c|}{{ \begin{tabular}[c]{@{}c@{}}\bf ADMM-\\ \bf ADMMCns\end{tabular}}}             & \multicolumn{2}{c|}{\begin{tabular}[c]{@{}c@{}}\bf FISTA-\\ \bf AGPCns\end{tabular}}                                    & \multicolumn{2}{c|}{\begin{tabular}[c]{@{}c@{}} \bf P-ADMM-\\ \bf ADMMCns\end{tabular}}                                  & \multicolumn{2}{c|}{\begin{tabular}[c]{@{}c@{}}\bf P-FISTA-\\ \bf APGCns\end{tabular}}                                  \\ \hline
\multicolumn{1}{|c|}{\bf K}                    & 
\begin{tabular}[c]{@{}c@{}}\bf Coef.\\ \bf Update\end{tabular} & \begin{tabular}[c]{@{}c@{}}\bf Dict.\\ \bf Update\end{tabular} & \begin{tabular}[c]{@{}c@{}}\bf Coef.\\ \bf Update\end{tabular} & \begin{tabular}[c]{@{}c@{}}\bf Dict.\\ \bf Update\end{tabular} & \begin{tabular}[c]{@{}c@{}}\bf Coef.\\ \bf Update\end{tabular} & \begin{tabular}[c]{@{}c@{}}\bf Dict.\\ \bf Update\end{tabular} & \begin{tabular}[c]{@{}c@{}} \bf Coef.\\ \bf Update\end{tabular} & \begin{tabular}[c]{@{}c@{}} \bf Dict.\\ \bf Update\end{tabular} \\ \hline
\multicolumn{1}{|c|}{}                     & 0.57                                                   & 0.67                                                   & 0.75                                                   & 0.31                                                   & 19 e-3                                                 & 26 e-3                                                 & 19 e-3                                                 & 15 e-3                                                 \\ \cline{2-9} 
\multicolumn{1}{|c|}{\multirow{-2}{*}{\bf 5}}  & \multicolumn{2}{c|}{1.24}                                                                                       & \multicolumn{2}{c|}{1.06}                                                                                       & \multicolumn{2}{c|}{45 e-3}                                                                                     & \multicolumn{2}{c|}{34 e-3}                                                                                     \\ \hline
\multicolumn{1}{|c|}{}                     & 1.11                                                   & 1.29                                                   & 1.49                                                   & 0.49                                                   & 34 e-3                                                 & 46 e-3                                                 & 37 e-3                                                 & 23 e-3                                                 \\ \cline{2-9} 
\multicolumn{1}{|c|}{\multirow{-2}{*}{\bf 10}} & \multicolumn{2}{c|}{2.4}                                                                                        & \multicolumn{2}{c|}{1.98}                                                                                       & \multicolumn{2}{c|}{80 e-3}                                                                                     & \multicolumn{2}{c|}{60 e-3}                                                                                     \\ \hline
\multicolumn{1}{|c|}{}                     & 2.18                                                   & 2.52                                                   & 2.98                                                   & 0.9                                                    & 72 e-3                                                 & 90 e-3                                                 & 73 e-3                                                 & 40 e-3                                                 \\ \cline{2-9} 
\multicolumn{1}{|c|}{\multirow{-2}{*}{\bf 20}} & \multicolumn{2}{c|}{4.7}                                                                                        & \multicolumn{2}{c|}{3.88}                                                                                       & \multicolumn{2}{c|}{162 e-3}                                                                                    & \multicolumn{2}{c|}{113 e-3}                                                                                    \\ \hline
\multicolumn{1}{|c|}{}                     & 4.34                                                   & 4.96                                                   & 5.96                                                   & 1.66                                                   & 184 e-3                                                & 220 e-3                                                & 157 e-3                                                & 74 e-3                                                 \\ \cline{2-9} 
\multicolumn{1}{|c|}{\multirow{-2}{*}{\bf 40}} & \multicolumn{2}{c|}{9.3}                                                                                        & \multicolumn{2}{c|}{7.62}                                                                                       & \multicolumn{2}{c|}{404 e-3}                                                                                    & \multicolumn{2}{c|}{231 e-3}                                                                                    \\ \hline
\end{tabular}}
\label{tab:mean_time_img}

\end{table}
\begin{table}[H]
\centering
\caption{Average time (in seconds) per iteration of each stage of the CDL algorithms varying the filter set size ($M$).}
\scalebox{0.9}{
\renewcommand{\arraystretch}{1.15}
\addtolength{\tabcolsep}{-3pt}
\begin{tabular}{c|c|c|c|c|c|c|c|c|}
\cline{2-9}
                                           & \multicolumn{2}{c|}{{ \begin{tabular}[c]{@{}c@{}}\bf ADMM-\\ \bf ADMMCns\end{tabular}}}             & \multicolumn{2}{c|}{\begin{tabular}[c]{@{}c@{}}\bf FISTA-\\ \bf AGPCns\end{tabular}}                                    & \multicolumn{2}{c|}{\begin{tabular}[c]{@{}c@{}} \bf P-ADMM-\\ \bf ADMMCns\end{tabular}}                                  & \multicolumn{2}{c|}{\begin{tabular}[c]{@{}c@{}}\bf P-FISTA-\\ \bf APGCns\end{tabular}}                                  \\ \hline
\multicolumn{1}{|c|}{\bf M}                    &
\begin{tabular}[c]{@{}c@{}}\bf Coef.\\ \bf Update\end{tabular} & \begin{tabular}[c]{@{}c@{}}\bf Dict.\\ \bf Update\end{tabular} & \begin{tabular}[c]{@{}c@{}}\bf Coef.\\ \bf Update\end{tabular} & \begin{tabular}[c]{@{}c@{}}\bf Dict.\\ \bf Update\end{tabular} & \begin{tabular}[c]{@{}c@{}}\bf Coef.\\ \bf Update\end{tabular} & \begin{tabular}[c]{@{}c@{}}\bf Dict.\\ \bf Update\end{tabular} & \begin{tabular}[c]{@{}c@{}} \bf Coef.\\ \bf Update\end{tabular} & \begin{tabular}[c]{@{}c@{}} \bf Dict.\\ \bf Update\end{tabular} \\ \hline
\multicolumn{1}{|c|}{}   & 0.22                                                   & 0.26                                                   & 0.31                                                   & 0.1                                                    & 10 e-3                                                 & 14 e-3                                                 & 10 e-3                                                 & 8 e-3                                                  \\ \cline{2-9} 
\multicolumn{1}{|c|}{\multirow{-2}{*}{\bf 4}}                     & \multicolumn{2}{c|}{0.48}                                                                                       & \multicolumn{2}{c|}{0.41}                                                                                       & \multicolumn{2}{c|}{24 e-3}                                                                                     & \multicolumn{2}{c|}{18 e-3}                                                                                     \\ \hline
\multicolumn{1}{|c|}{}  & 0.96                                                   & 1.14                                                   & 1.32                                                   & 0.4                                                    & 31 e-3                                                 & 41 e-3                                                 & 32 e-3                                                 & 20 e-3                                                 \\ \cline{2-9} 
\multicolumn{1}{|c|}{\multirow{-2}{*}{\bf 16}}                     & \multicolumn{2}{c|}{2.1}                                                                                        & \multicolumn{2}{c|}{1.72}                                                                                       & \multicolumn{2}{c|}{72 e-3}                                                                                     & \multicolumn{2}{c|}{52 e-3}                                                                                     \\ \hline
\multicolumn{1}{|c|}{}  & 2.18                                                   & 2.52                                                   & 2.98                                                   & 0.9                                                    & 73 e-3                                                 & 90 e-3                                                 & 73 e-3                                                 & 40 e-3                                                 \\ \cline{2-9} 
\multicolumn{1}{|c|}{\multirow{-2}{*}{\bf 36}}                     & \multicolumn{2}{c|}{4.7}                                                                                        & \multicolumn{2}{c|}{3.88}                                                                                       & \multicolumn{2}{c|}{163 e-3}                                                                                    & \multicolumn{2}{c|}{113 e-3}                                                                                    \\ \hline
\multicolumn{1}{|c|}{}  & 3.8                                                    & 4.41                                                   & 5.2                                                    & 1.57                                                   & 166 e-3                                                & 204 e-3                                                & 139 e-3                                                & 68 e-3                                                 \\ \cline{2-9} 
\multicolumn{1}{|c|}{\multirow{-2}{*}{\bf 64}}                     & \multicolumn{2}{c|}{8.21}                                                                                       & \multicolumn{2}{c|}{6.77}                                                                                       & \multicolumn{2}{c|}{370 e-3}                                                                                    & \multicolumn{2}{c|}{207 e-3}                                                                                    \\ \hline
\end{tabular}}
\label{tab:mean_time_filter}
\end{table}

In addition, we can see in the Figure \ref{fig:mean_time_filter_img}, and  Tables \ref{tab:mean_time_img} and \ref{tab:mean_time_filter} that, regardless of the number of training images or number of filters, our regular and parallel FISTA-APGCns algorithms exhibit a lower average time per iteration than regular and parallel ADMM-ADMMCns algorithms respectively. The speedup factor of our CDL implementations (1.25x to 1.75x) is directly proportional to the training  set size and filter set size, i.e. for a larger set a higher speedup factor. If we focus specifically on dictionary update
stage (see Dict. Update in the Tables \ref{tab:mean_time_img} and \ref{tab:mean_time_filter}), we can notice that our proposed method (APG consensus) w.r.t. its respective counterpart (ADMM consensus) attains superior speedup (2x to 3x) with a similar proportional increase.
 
Although the CDL problem has an $\ell_1$ penalty term that can intrinsically avoid over-fitting, we additionally analyse the generalization of the dictionaries, all with 36 filters, during training process and in a denoising application. As the type of implementation (regular or parallel) does not  affect the dictionary generalization,  we evaluate the performance on the dictionaries estimated from the regular implementations.
For both cases, we used a group of 20 test images of the MIRFFLICKR-IM dataset, with dimensional characteristics equal to those of the training set. 

In order to check the generalization during training, we extract a dictionary each 50 iterations and test it on the convolutional Basis Pursuit DeNoising (CBPDN) problem:
\begin{equation}
\argmin _{ \{ {x}_{m}\}  }  \frac{1}{2}  \norm{\fsum_{m}   {d}_m * {x}_{m} - {s}}_2^2 
+ \lambda \fsum_{m} \|{x}_{m} \|_1, 
\label{CBPDN-1}
\end{equation}
where $d_m$ is a given dictionary and $x_m$ represents the coefficient maps that encode an image $s$. To standardize results, the problem (\ref{CBPDN-1}) is computed via a CSC algorithm of the SPORCO library \cite{sporco_brendt} with a sparsity parameter $\lambda = 0.1$ and
200 fixed iterations.

\begin{figure}[H]
\centering
\includegraphics[width=0.28\textwidth]{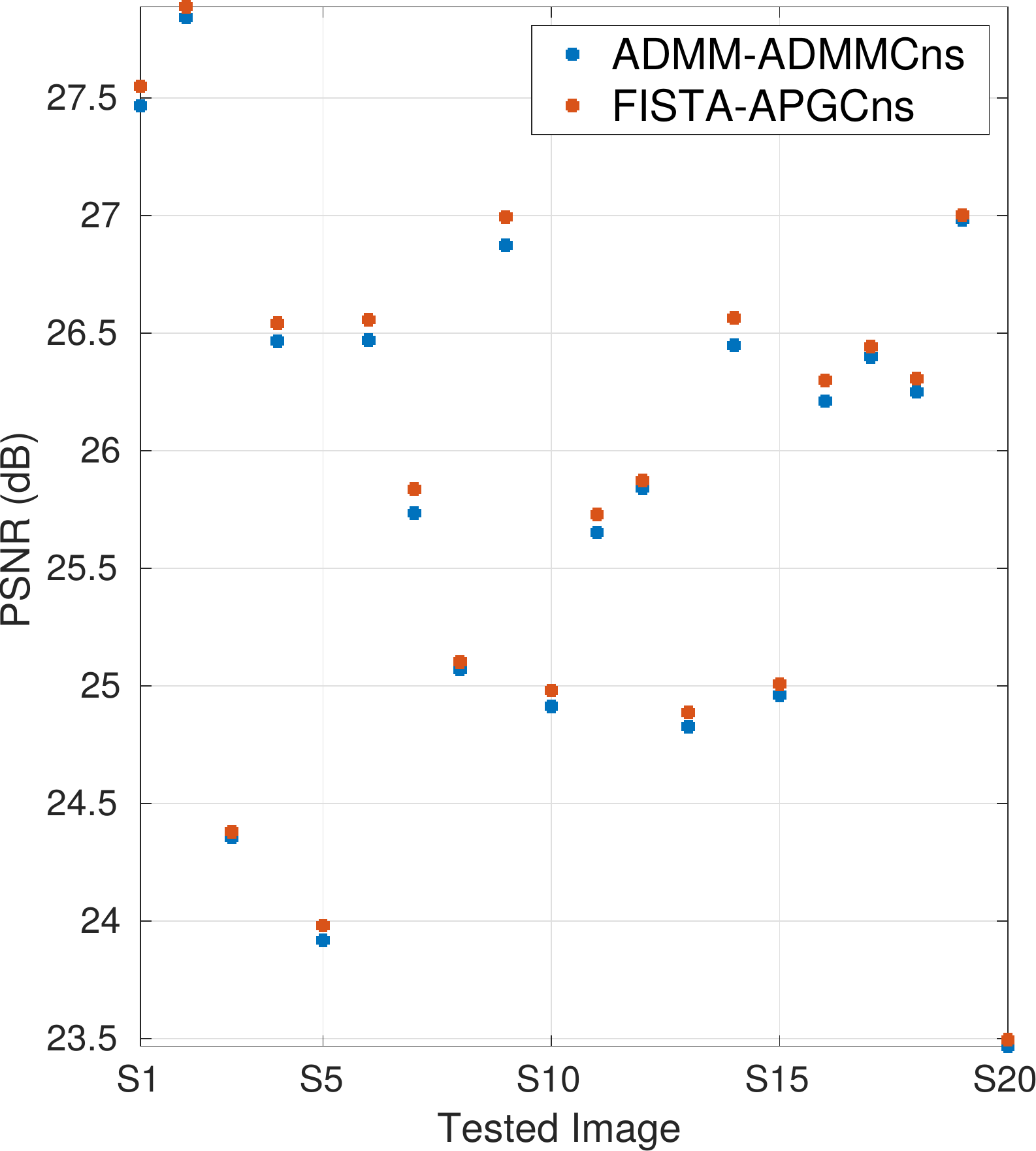}
\hspace{5mm}
\includegraphics[width=0.265\textwidth]{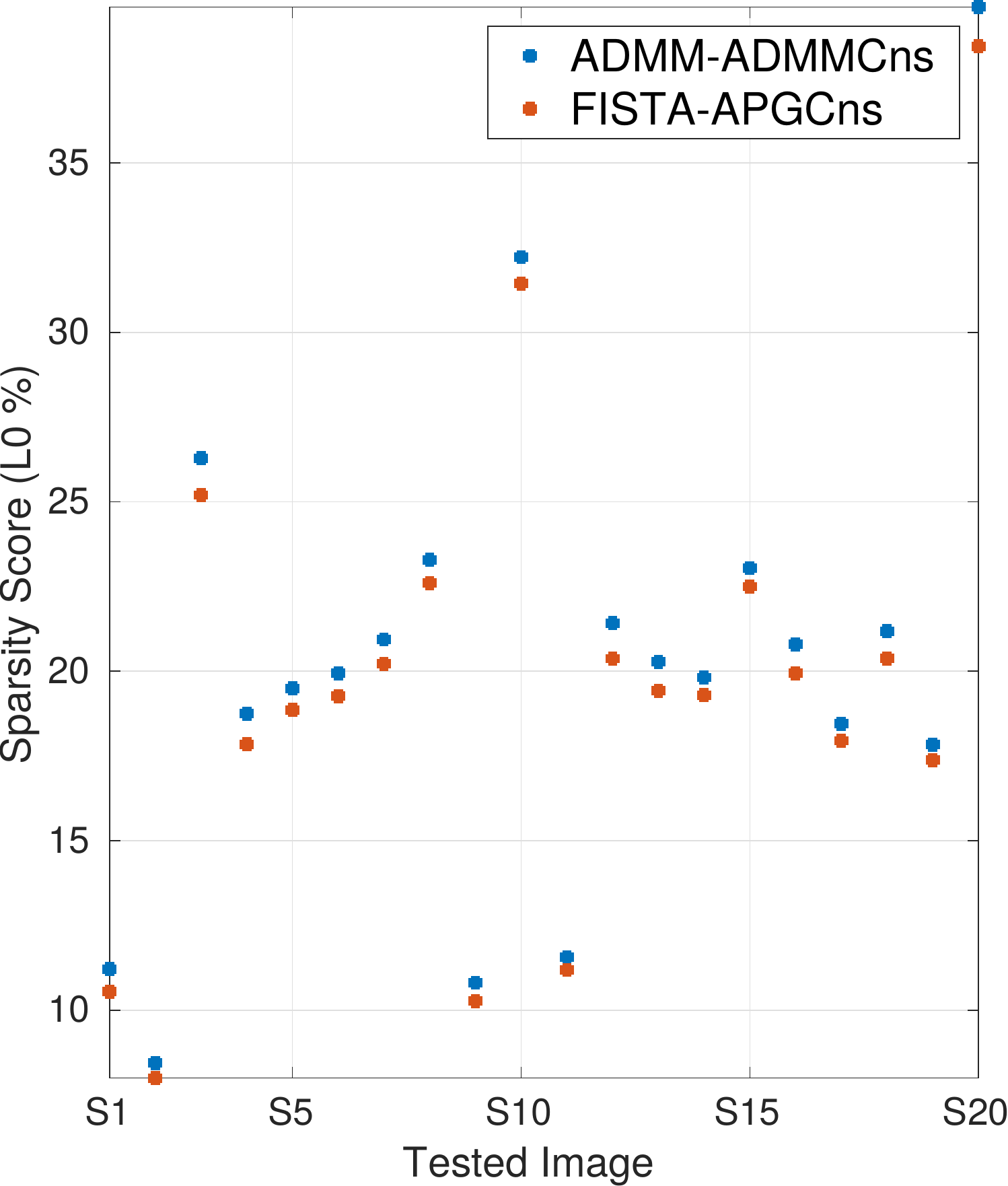} 
\caption{Denoising comparisons for a pair of dictionaries learned from 40 training images.}
\label{fig:PSNR_result}
\end{figure}

For the denoising application, each tested image is corrupted with Additive White Gaussian noise (AWGN) of level $\sigma = 0.1$. As reconstruction metrics, we consider the PSNR  and a sparsity measure defined as $100\cdot  \sum_m ||x_m||_0 / N$, where $N$ is the number of pixels of the tested image. In this second case, the CBPDN problem (\ref{CBPDN-1}) has an optimal sparsity parameter $\lambda$ that maximizes the PSNR score. 
Because of this, we found an optimal parameter per image for the ADMM-ADMMCns dictionary and used it for both algorithms (ADMM-ADMMCns and FISTA-APGCns); such results are illustrated in Figure \ref{fig:PSNR_result}.  

\begin{figure}[H]
\centering
\includegraphics[width=0.29\textwidth]{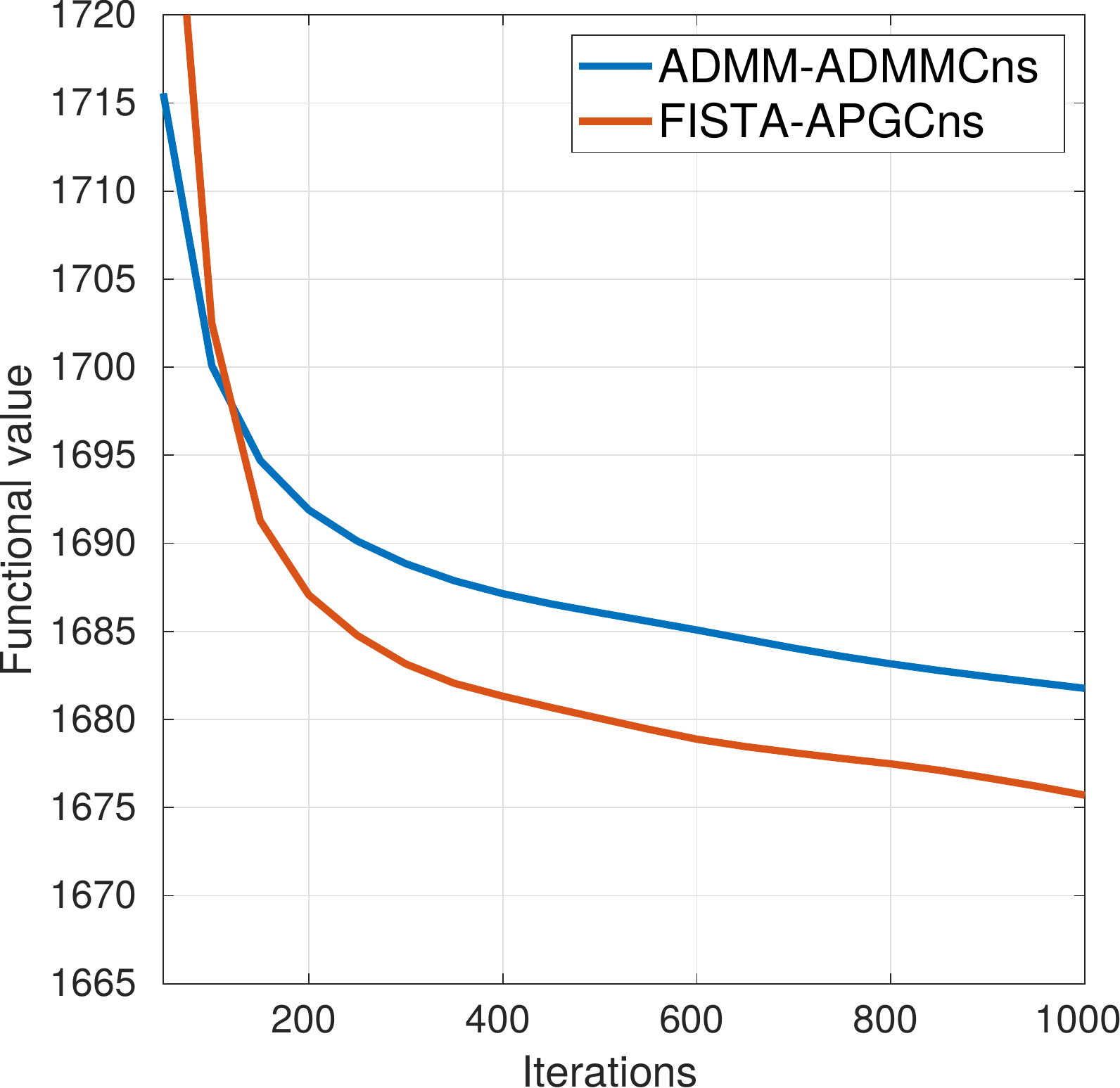}   
\hspace{5mm}
\includegraphics[width=0.29\textwidth]{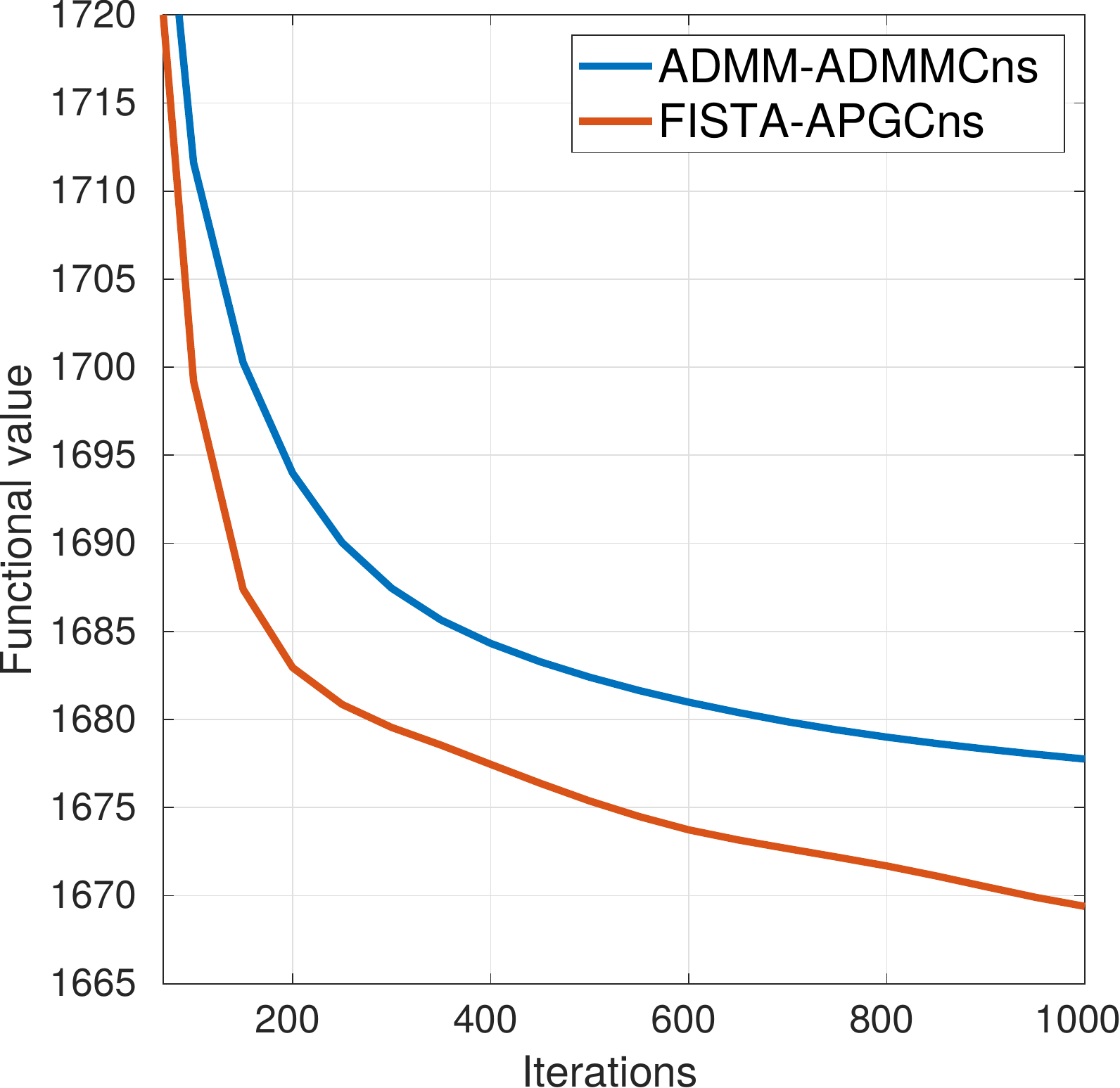} 
\caption{ Progress of each final functional value of CBPDN using the test set and each partial dictionary obtained when training for  20 images (Left) and 40 images (Right)}
\label{fig:test_cost_iter_20img}
\end{figure}

It can be observed in Figure \ref{fig:test_cost_iter_20img} that the dictionaries learned with a larger training set improves performance in the test stage regardless of the minimizer algorithm.
Moreover, for the same number of iterations the dictionaries computed via our proposed algorithm can induce better sparsity or fidelity.

\vspace{-5mm}
Note that for the denoising task (see the Figure \ref{fig:PSNR_result}), the dictionary estimated by our algorithm yields slightly better PSNR scores with fewer non-zero components (lower sparsity measure). However, in practice, the performance of both dictionaries can be considered equivalent.

\subsection{\bfseries Results in anomaly detection task}

Sparse coding and its convolutional counterpart have shown to provide remarkable results in the anomaly detection tasks \cite{carrera2015detecting,pilastre2020anomaly,takeishi2014anomaly,pilastre2020}, especially for mixed continuous and discrete data. In particular, \cite{pilastre2020anomaly} recently introduced a consensus convolutional formulation\footnote{The anomaly detection problem  (\ref{eqn:C-ADDICT-1}) was simplified for illustrative purposes, check \cite{pilastre2020anomaly} for full details.} (\ref{eqn:C-ADDICT-1}) for identifying univariate and multivariate anomalies $e_{p}$ in mixed data $s_{p}$ with $P$ time-series. 
\begin{eqnarray}
\mathop {\mathrm {arg\,min\kern 0pt}} _{  \{{x}_{m,p}\} \{ {e}_{p}\} }  \frac{1}{2}  \fsum_{p} \norm{\fsum_{m}  {d}_{m,p} * {x}_{m,p} + {e}_{p} - {s}_{p}}_2^2
+ \lambda\fsum_{p,m} \|{x}_{m,p} \|_1 +  \beta \fsum_{p} \|{e}_{p}\|_2 
 \label{eqn:C-ADDICT-1} 
\\ \notag \\
\text{\vspace{15mm}s.t.\hspace{2.5mm}} {x}_{m,1} =  {x}_{m,2}  = \cdots =  {x}_{m,P} \notag
\end{eqnarray}
In this formulation, the sparse representations $x_{m,p}$ and the agreement constraint enforce a joint and uniform activation of filters $d_{m,p}$ in order to capture correlations across time-series.

For this second case,  we evaluate the following algorithms:
\begin{itemize}
\item \textbf{ADMMCns}: Anomaly detection algorithm based on a ADMM-consensus approach proposed in \cite{pilastre2020anomaly} with the name of $\text{C-ADDICT}$ for solving (\ref{eqn:C-ADDICT-1}).

\item \textbf{APGCns}:  Our APG-consensus algorithm for the anomaly detection problem (\ref{eqn:C-ADDICT-1}).
\end{itemize}

To not incur redundant information\footnote{We already showed in the first part that the regular and parallel AGP consensus algorithms provide a better computational benefit than the respective ADMM counterparts.}, the algorithms listed above are regular MATLAB implementations. 
The considered dataset, Seattle Burke Gilman Trail dataset \cite{seattle_2020}, is a collection of time series (acquired by different sensors) from 2014 to 2019, which contains the pedestrian and cyclist counts  for the south and north directions of travel.  
Since both anomaly detection algorithms need a pre-trained dictionary, we learned a single dictionary of 200 filters of length 100 per time-series using the CDL algorithm proposed in  \cite{pilastre2020anomaly} from a portion of the dataset (2014 to 2015), where the number of anomalous accounts is minimal.  

\vspace{-2.5mm}
\begin{figure}[H]
\begin{center}
\includegraphics[width=0.29\textwidth]{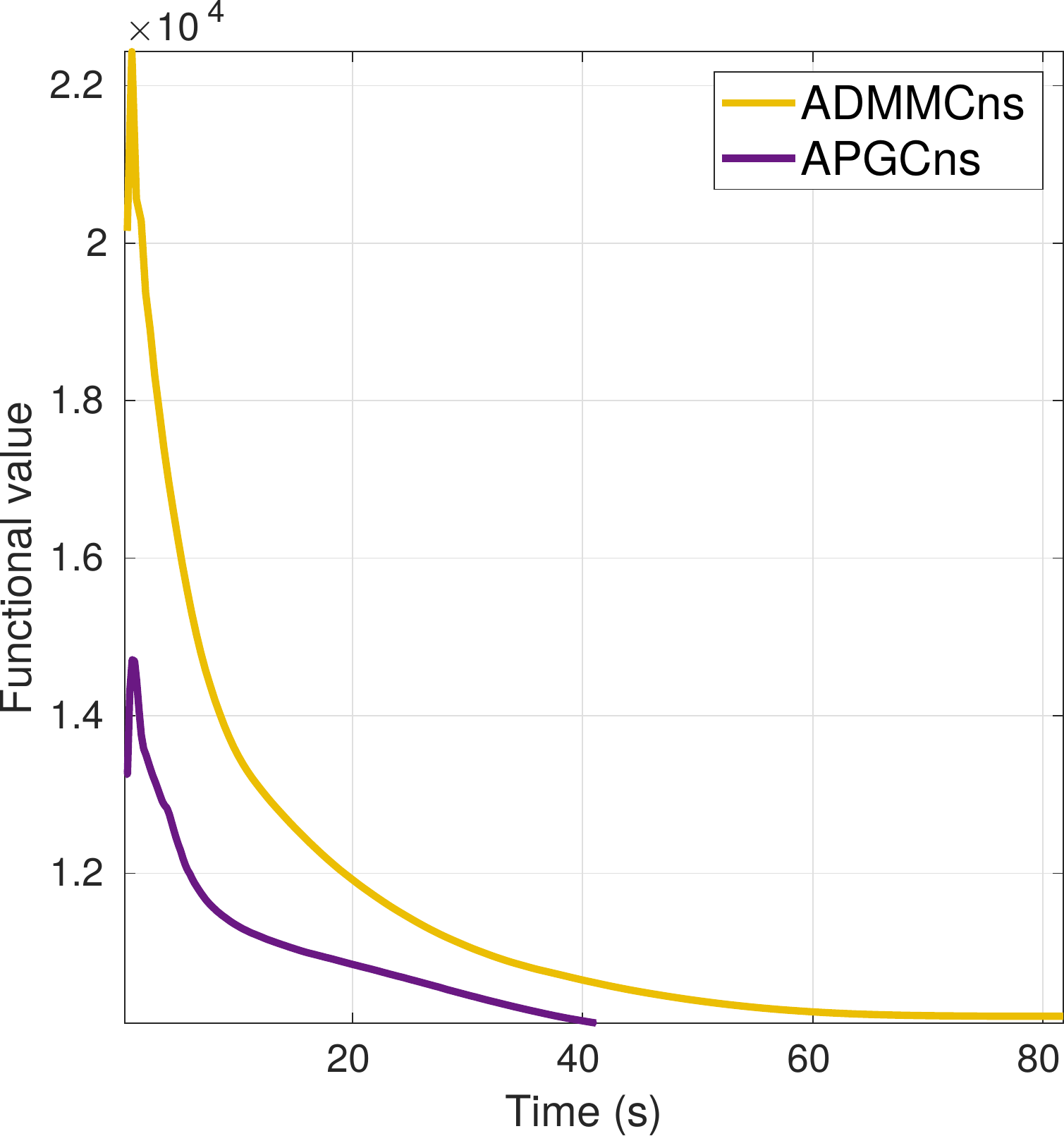} 
\hspace{5mm}
\includegraphics[width=0.29\textwidth]{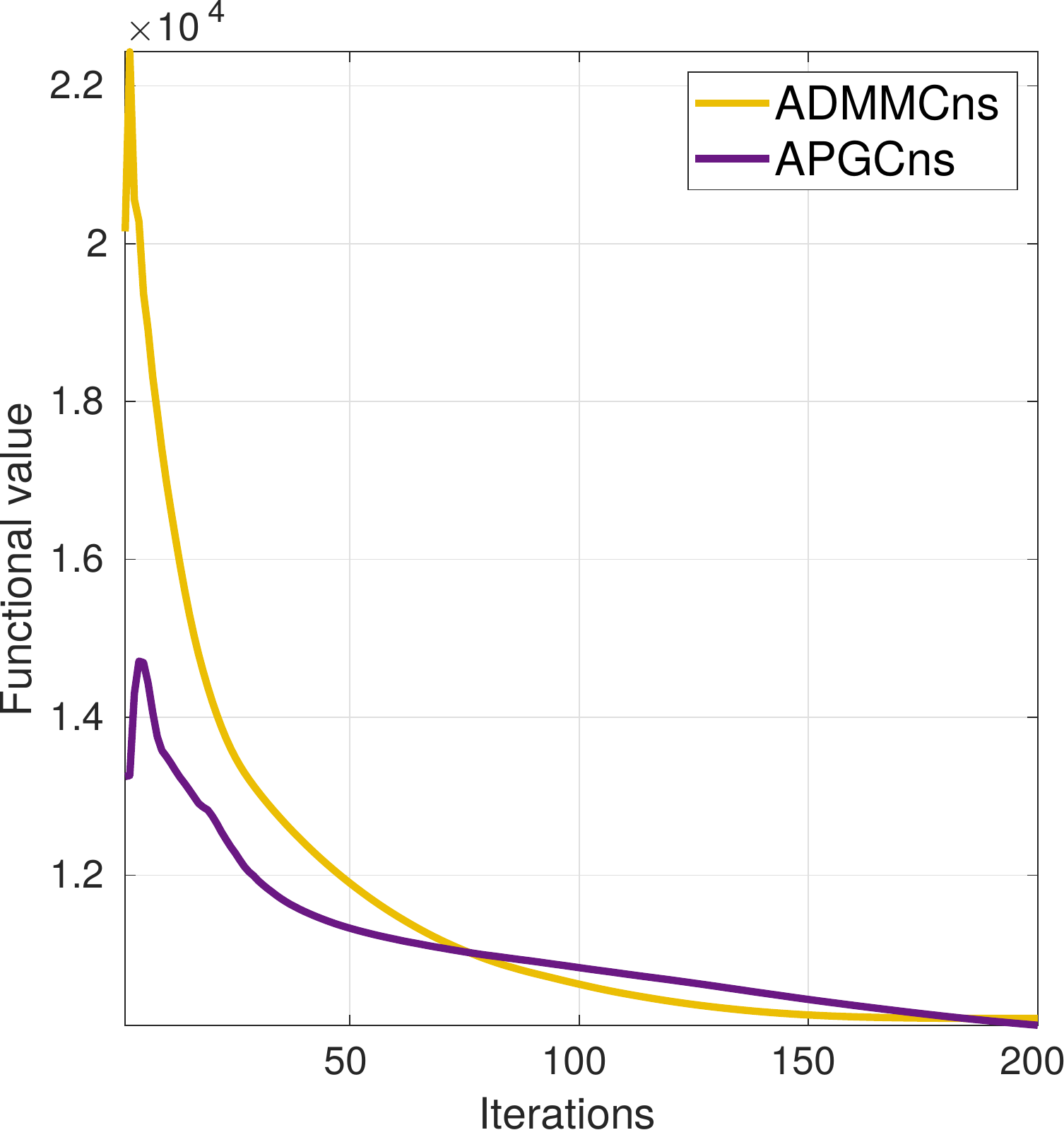}     
\end{center}
\vspace{-2mm}
\caption{A comparison of the functional value decay with respect to run-time and iterations.}
\label{fig:cost_time_time_anomalies}
\end{figure}
\vspace{-3.5mm}
\begin{figure}[H]
\centering
\includegraphics[width=0.6\textwidth]{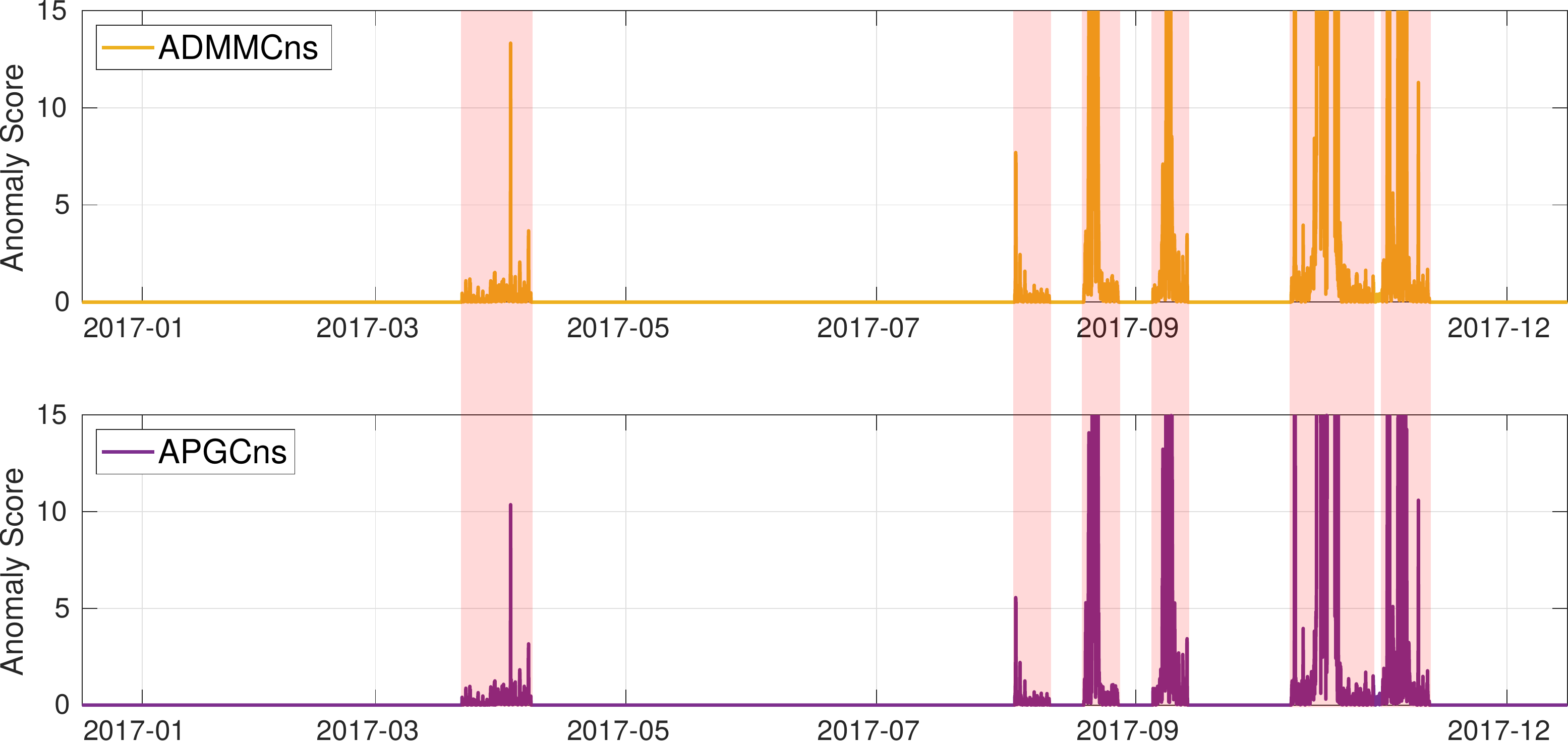} 
\caption{Anomalies detected marked by red background, where anomaly score is $\ell_2$ norm of the anomaly vector $e_p$ across time-series.}
\label{fig:anomalies}
\end{figure}

In  Figures \ref{fig:cost_time_time_anomalies} and \ref{fig:anomalies}, we report the performance comparisons of the anomaly detection algorithms on the data of the year 2017. Similarly to the CDL related experiments, our APGCns has a better convergence with respect to run-time. Although 
both algorithms present distinct evaluations of functional value per iteration, these achieve a similar  point of convergence. As exported, we can observed in the Figure \ref{fig:anomalies}
that a same pattern of anomalous accounts, highlighted by red background, is detected since we only modified the minimizer algorithm of  (\ref{eqn:C-ADDICT-1})  which should not alter the detection if the same convergence point is achieved.

\section{\bfseries Conclusions}
\label{Sect:conclusions}
This article has introduced an efficient and generic consensus method based on the accelerated proximal gradient (APG) for solving optimization problems of the form  $min_x\sum_{i} f_i({x}) + g (x)$. Our derivation exploits a direct consensus formulation which avoids the use of the common auxiliary variable (usually found in ADMM-based approaches). Interestingly, the final structure of our APG-based consensus resembles that of ADMM-based but with the usual APG benefits, i.e. simplicity and low complexity per iteration, which is reflected in its computational performance.

For instance, in the experiments related to the CDL problem, our APG consensus method has shown to significantly outperform the ADMM counterpart in terms of convergence and run-time without any loss of quality (generalization) in the estimated dictionaries. Furthermore, its general formulation allowed us: (i)  To easily derive an adequate step-size in order to have a self-adjusting algorithm. (ii)
To apply in other optimization problems such as in the case of anomaly detection, in which we additionally reinforced the computational efficiency of the proposed method. All the implementations based on our method, presented in this manuscript, are available in \cite{apg_gustavo}.

\ifCLASSOPTIONcaptionsoff
  \newpage
\fi

\bibliographystyle{spmpsci}      % mathematics and physical sciences
\bibliography{references.bib}

% that's all folks
\end{document}